\documentclass[acmsmall,natbib,screen]{acmart}

\usepackage{ragged2e}
\usepackage{textcomp}
\usepackage{xcolor}
\usepackage{graphicx}
\usepackage{url}
\usepackage{xspace}
\usepackage{latexsym}
\usepackage{paralist}
\usepackage{amsmath}
\usepackage{amsfonts}
\usepackage{multirow}
\usepackage{subfigure}
\usepackage{booktabs}
\usepackage{color}
\usepackage{threeparttable}
\usepackage{xAlgorithm}
\usepackage{booktabs} % For formal tables
\usepackage{makecell}

\usepackage{amsmath} % used for boldsymbol.
\renewcommand{\vec}[1]{\boldsymbol{#1}} % Uncomment for BOLD vectors.

\newcommand{\fb}{\texttt{DBP-FB}\xspace}
\newcommand{\our}{\textsf{CEAFF}\xspace}
\newcommand{\ourc}{\textsf{CEAFF-Coh}\xspace}
\newcommand{\oure}{\textsf{CEAFF-Excl}\xspace}
\newcommand{\avg}{\textsf{AVG}\xspace}
\newcommand{\oursm}{\textsf{CEAFF-SM}\xspace}

\newcommand{\rsn}{\textsf{RSNs}\xspace}
\newcommand{\mtranse}{\textsf{MTransE}\xspace}

\newcommand{\itranse}{\textsf{ITransE}\xspace}
\newcommand{\jape}{\textsf{JAPE}\xspace}
\newcommand{\bootea}{\textsf{BootEA}\xspace}
\newcommand{\gcn}{\textsf{GCN-Align}\xspace}

\newcommand{\gm}{\textsf{GM-Align}\xspace}

\newcommand{\mc}{\textsf{MuGNN}\xspace}
\newcommand{\na}{\textsf{NAEA}\xspace}
\newcommand{\rd}{\textsf{RDGCN}\xspace}
\newcommand{\kecg}{\textsf{KECG}\xspace}
\newcommand{\te}{\textsf{TransEdge}\xspace}
\newcommand{\hgcn}{\textsf{HGCN}\xspace}
\newcommand{\hman}{\textsf{HMAN}\xspace}
\newcommand{\cea}{\textsf{CEA}\xspace}
\newcommand{\dat}{\textsf{DAT}\xspace}
\newcommand{\transe}{\textsf{TransE}\xspace}
\newcommand{\paris}{\textsf{PARIS}\xspace}

\newcommand{\ali}{\textsf{AliNet}\xspace}
\newcommand{\gmehd}{\textsf{GM-EHD-JEA}\xspace}
\newcommand{\mraea}{\textsf{MRAEA}\xspace}

\newcommand{\lev}{\textsf{Lev}\xspace}

\newcommand{\dbps}{\texttt{DBP15K}\xspace}
\newcommand{\dbpsz}{$\texttt{DBP15K}_\texttt{ZH-EN}$\xspace}
\newcommand{\dbpsj}{$\texttt{DBP15K}_\texttt{JA-EN}$\xspace}
\newcommand{\dbpsf}{$\texttt{DBP15K}_\texttt{FR-EN}$\xspace}

\newcommand{\srp}{\texttt{SRPRS}\xspace}
\newcommand{\srpf}{$\texttt{SRPRS}_\texttt{EN-FR}$\xspace}
\newcommand{\srpd}{$\texttt{SRPRS}_\texttt{EN-DE}$\xspace}
\newcommand{\srpw}{$\texttt{SRPRS}_\texttt{DBP-WD}$\xspace}
\newcommand{\srpy}{$\texttt{SRPRS}_\texttt{DBP-YG}$\xspace}

\newcommand{\sota}{state-of-the-art\xspace}

\newtheorem{example}{Example}

\newcommand{\myurl}[1]{\url{#1}}
\newcommand{\myfootnote}[1]{\footnote{\small #1}}
\newcommand{\goodgap}{\hspace{\subfigtopskip}\hspace{\subfigbottomskip}}

\newcommand{\myparagraph}[1]{\vspace{1ex}\noindent\textbf{#1.}\hspace{1em}}

\SetAlFnt{\small}
\SetAlCapFnt{\small}
\SetAlCapNameFnt{\small}
\SetAlCapHSkip{0pt}
\IncMargin{-\parindent}

\AtBeginDocument{%
  \providecommand\BibTeX{{%
    \normalfont B\kern-0.5em{\scshape i\kern-0.25em b}\kern-0.8em\TeX}}}

\setcopyright{acmcopyright}
\copyrightyear{2021}
\acmYear{2021}
\acmDOI{10.1145/1122445.1122456}

\begin{document}

\title[RL-based Collective Entity Alignment with Adaptive Features]{Reinforcement Learning based Collective Entity Alignment with Adaptive Features}


\author{Weixin Zeng}
\email{zengweixin13@nudt.edu.cn}
\author{Xiang Zhao}
\authornote{Corresponding author.}
\email{xiangzhao@nudt.edu.cn}
\author{Jiuyang Tang}
\email{jiuyang\_tang@nudt.edu.cn}
\affiliation{%
  \institution{Science and Technology on Information Systems Engineering Laboratory, National University of Defense Technology}
  \city{Changsha}
  \country{China}
}
%  \streetaddress{P.O. Box 1212}

\author{Xuemin Lin}
\email{lxue@cse.unsw.edu.au}
\affiliation{%
  \institution{The University of New South Wales}
  \city{Sydney}
  \country{Australia}
}

\author{Paul Groth}
\email{p.groth@uva.nl}
\affiliation{%
	\institution{University of Amsterdam}
	\city{Amsterdam}
	\country{The Netherlands}
}

\thanks{This work was partially supported by 
	Ministry of Science and Technology of China under grant No. 2020AAA0108800,
	NSFC under grants Nos. 61872446 and 71971212, 
	NSF of Hunan Province under grant No. 2019JJ20024,  
	and Postgraduate Scientific Research Innovation Project of Hunan Province under grant No. CX20190033. 
	All content represents the opinion of the authors, which is not necessarily shared or endorsed by their respective employers and/or sponsors.}

\renewcommand{\shortauthors}{W. Zeng et al.}

\begin{abstract}
  Entity alignment (EA) is the task of identifying the entities that refer to the same real-world object but are located in different knowledge graphs (KGs). For entities to be aligned, existing EA solutions treat them separately and generate alignment results as ranked lists of entities on the other side. Nevertheless, this decision-making paradigm fails to take into account the interdependence among entities. Although some recent efforts mitigate this issue by imposing the 1-to-1 constraint on the alignment process, they still cannot adequately model the underlying interdependence and the results tend to be sub-optimal. 
  
  To fill in this gap, in this work, we delve into the dynamics of the decision-making process, and offer a reinforcement learning (RL) based model to align entities collectively. Under the RL framework, we devise the coherence and exclusiveness constraints to characterize the interdependence and restrict collective alignment. Additionally, to generate more precise inputs to the RL framework, we employ representative features to capture different aspects of the similarity between entities in heterogeneous KGs, which are integrated by an adaptive feature fusion strategy. Our proposal is evaluated on both cross-lingual and mono-lingual EA benchmarks and compared against state-of-the-art solutions. The empirical results verify its effectiveness and superiority.
\end{abstract}





\maketitle

\section{Introduction}
Knowledge graphs (KGs) play a pivotal role in tasks such as information retrieval~\cite{DBLP:conf/www/XiongPC17}, question answering~\cite{DBLP:conf/naacl/HixonCH15} and recommendation systems~\cite{DBLP:conf/www/0003W0HC19}.
Although many KGs have been constructed over recent years, none of them can guarantee \emph{full coverage}~\cite{DBLP:journals/semweb/Paulheim17}. 
Indeed, KGs often contain complementary information, which motivates the task of merging KGs.  
%Having been constructed from different sources, nonetheless, these heterogeneous KGs usually contain complementary contents, which makes it alluring to exploit the integration of multiple KGs.

To incorporate the knowledge from a \emph{target} KG into a \emph{source} KG, an important step is to align entities between them.
To this end, the task of \emph{entity alignment} (EA) is proposed, which aims to discover entities that have the same meaning but belong to different original KGs. 
To handle the task, \sota EA methods~\cite{DBLP:conf/acl/CaoLLLLC19,DBLP:conf/www/PeiYHZ19,DBLP:conf/ijcai/SunHZQ18,DBLP:conf/ijcai/ZhuZ0TG19} assume that equivalent entities in different KGs have similar neighborhood structures. 
As a consequence, they first use representation learning technologies (e.g., \transe~\cite{DBLP:conf/ijcai/ChenTYZ17} and graph convolutional network (GCN)~\cite{DBLP:conf/emnlp/WangLLZ18}) to capture structural features of KGs; that is, they map entities into data points in a low-dimensional feature space, where pair-wise similarity can be easily evaluated between the data points. 
Then, to determine the alignment result, they treat entities \emph{independently} and retrieve a ranked list of target entities for each source entity according to the pair-wise similarities, where the top-ranked target entity is selected as the predicted match to the source entity in question. 

%\textcolor{blue}{
%Nevertheless, we observe that \sota EA methods cannot effectively tackle the multi-feature fusion and alignment decision-making problems during the alignment process. 
%%, and there is still a large room for improvement.  
%Next, we introduce these deficiencies and our solutions.}

%\myparagraph{\textcolor{blue}{Multi-feature Fusion}}
Nevertheless, merely using the KG structure, in most cases, cannot guarantee satisfactory alignment results~\cite{DBLP:conf/ijcai/ChenTCSZ18}. 
Consequently, recent methods exploit multiple types of features, e.g., attributes~\cite{DBLP:conf/aaai/TrisedyaQZ19,DBLP:conf/emnlp/WangLLZ18}, entity description~\cite{DBLP:conf/emnlp/YangZSLLS19,DBLP:conf/ijcai/ChenTCSZ18}, entity names~\cite{DBLP:conf/emnlp/WuLFWZ19,DBLP:conf/ijcai/WuLF0Y019}, to provide a more comprehensive view for alignment.
To fuse different features, these methods first calculate pair-wise similarity scores within each feature-specific space, and then combine these scores to generate the final similarity score.
%some approaches directly learn and combine feature-specific embeddings to generate an aggregated embedding for each entity, which is then used to calculate pair-wise similarity between entities~\cite{IJCAI19,ACL19,DBLP:conf/ijcai/WuLF0Y019}. 
%Nevertheless, these representation-level fusion techniques might fail to maintain
%the characteristics of the original features, e.g., two entities might be extremely similar in feature-specific embedding spaces, while placed distantly in the unified representation space.
%Meanwhile, some methods~\cite{DBLP:conf/emnlp/WangLLZ18,DBLP:conf/semweb/SunHL17} adopt the outcome-level fusion strategy that operates on the intermediate outcomes--feature-specific similarity matrices. 
%They first calculate pairwise similarity scores within each feature-specific space, and then combine these scores to generate the final similarity score. 
However, they manually assign the weights of features, which can be impractical when the number of features increases, or the importance of certain features varies greatly under different settings.
%Therefore, it is of significance to dynamically determine the weight of each feature.}

%Moreover, as mentioned before, a more accurate similarity matrix can also lead to better alignment results. 
%we also strive to generate a more accurate similarity matrix. 
%To this end, 
In response, we first exploit the structural, semantic, string-level features to capture different aspects of the similarity between the entities in source and target KGs. 
Then, to effectively aggregate different features, we devise an adaptive feature fusion strategy to fuse the feature-specific similarity matrices.
The similarity matrices are calculated using the nature-inspired Bray-Curtis dissimilarity~\cite{bray1957ordination}, a widely used measure to quantify the compositional difference between two coenoses, which can better capture the similarity between entities than commonly used distance measures, e.g., Manhattan distance~\cite{DBLP:conf/ijcai/WuLF0Y019,DBLP:conf/emnlp/WuLFWZ19,DBLP:conf/emnlp/WangLLZ18}, Euclidean distance~\cite{DBLP:conf/ijcai/ChenTYZ17,DBLP:conf/ijcai/ZhuXLS17,DBLP:conf/emnlp/YangZSLLS19}, and cosine similarity~\cite{DBLP:conf/semweb/SunHHCGQ19,DBLP:conf/semweb/SunHL17,DBLP:conf/ijcai/SunHZQ18}.
%Additionally, to consolidate comprehensive information and obtain the fused similarity matrix, we exploit multiple representative features to encode different aspects of the similarity between the entities in source and target KGs, and conceive an adaptive feature fusion strategy to effectively fuse different features at the \emph{outcome-level}, i.e., feature-specific similarity matrices calculated using the nature-inspired Bray-Curtis dissimilarity~\cite{bray1957ordination}, which is a widely used measure to quantify the compositional difference between two coenoses.
%\myparagraph{\textcolor{blue}{Alignment Decision-making}}
After obtaining the similarity matrix that encodes alignment signals from different features, the next crucial step is to make alignment decisions based on this matrix. 
As mentioned above, current methods adopt the independent decision-making strategy to generate alignment results, which is illustrated in Example~\ref{exa:motivate}.
%This paradigm of \emph{independent decision-making} is illustrated in Example~\ref{exa:motivate}.

\begin{figure*}[htbp]
	\centering
	\subfigure[Input KGs]{
		\begin{minipage}[t]{0.5\linewidth}
			\centering
			\includegraphics[width=5cm]{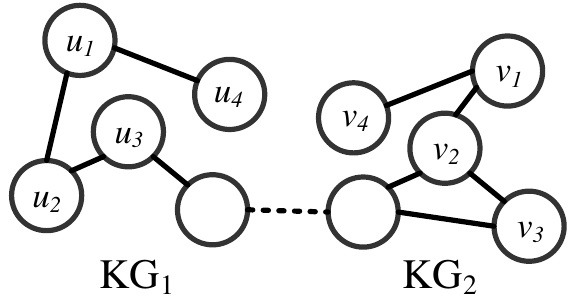}
		\end{minipage}%
	}%
	\subfigure[Similarity Matrix]{
		\begin{minipage}[t]{0.5\linewidth}
			\centering
			\includegraphics[width=3cm]{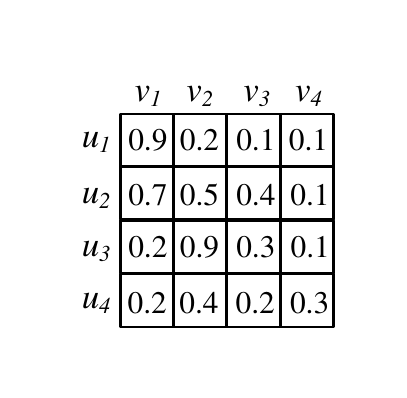}
		\end{minipage}%
	}
	\subfigure[Independent]{
		\begin{minipage}[t]{0.25\linewidth}
			\centering
			\includegraphics[width=2.5cm]{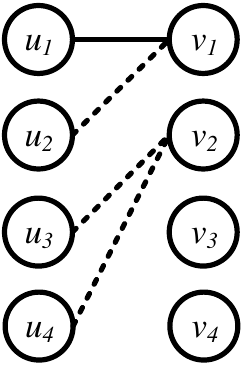}
		\end{minipage}%
	}
	\subfigure[1-to-1 Constrained]{
		\begin{minipage}[t]{0.25\linewidth}
			\centering
			\includegraphics[width=2.5cm]{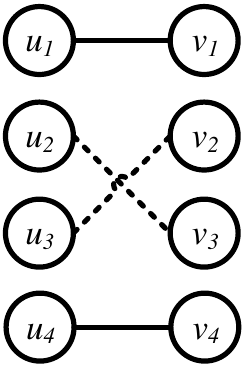}
		\end{minipage}%
	}
	\subfigure[Coherent + Exclusive]{
		\begin{minipage}[t]{0.25\linewidth}
			\centering
			\includegraphics[width=3.25cm]{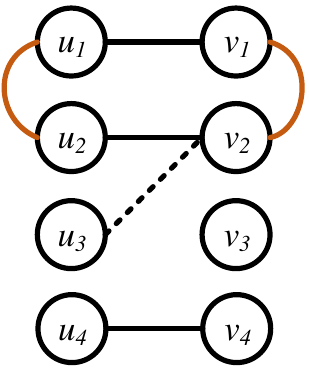}
		\end{minipage}%
	}
	\caption{An example of EA and different strategies.
		(a): Input KGs, where the nodes (resp. lines) depict entities (resp. relations); %The dashed lines denote seeInput KGs, where the nodes (resp. lines) depict entities (resp. relations)d entity pairs (training set);
		(b): Similarity matrix of all-pair entities; %between source and target entities;
		(c)---(e): Alignment produced by different strategies, where the entities connected by solid lines are correct matches, whereas those connected by dashed lines are erroneous matches, and the colored lines denote related entities in the same KG.}\label{fig:1}
\end{figure*}

\begin{example} \label{exa:motivate}
	In Figure~\ref{fig:1}(a) are two KGs ($KG_1$ and $KG_2$), where the dashed lines indicate known alignment (i.e., seeds). The target entities $v_1$, $v_2$, $v_3$, and $v_4$ in $KG_2$ are to be aligned, respectively, to the source entities $u_1$, $u_2$, $u_3$, and $u_4$ in $KG_1$ (i.e., ground truth). 
	Assume that we now arrive at some similarity matrix in Figure~\ref{fig:1}(b) for the entities in question, where greater values indicate higher similarities.  %Note that the entities with the identical indexes are equivalent, respectively.
	Following the aforementioned independent decision-making paradigm, %For state-of-the-art EA methods, the alignment decisions are made \emph{independently}. Concretely, regarding
	for the source entity $u_1$, as the target entity $v_1$ is of the highest similarity score, $(u_1, v_1)$ is predicted as a match; similarly, $(u_2, v_1)$, $(u_3, v_2)$ and  $(u_4, v_2)$ are predicted to be equivalent. % by independent EA solutions.
	Compared with the ground truth, the former is correct, and yet the latter three are erroneous.
	%; in fact, same conclusion can be drawn when the matrix in (c) or (d) is used. %, for (b), (c) and (d).
	%	These mismatches, however, can be avoided by exerting a simple \textbf{collective} alignment constraint, e.g., different source entities cannot match the same target entity and the source entity with higher similarity score can keep the match.
	%	In this case,
\end{example}

Example~\ref{exa:motivate} shows that the independent decision-making strategy fails to produce satisfactory alignment results, as it neglects the interdependence among the decision-making for each source entity. 
%%To reduce the erroneous matches produced by the decision-making process,  
%%a common solution is to devise more advanced representation learning models to increase the quality of structural embeddings, which in turn generates a more accurate similarity matrix~\cite{DBLP:conf/aaai/SunW0CDZQ20,DBLP:conf/wsdm/MaoWXLW20,DBLP:conf/ijcai/WuLF0Y019}.
%%\textcolor{red}{ 
%%Nevertheless, the effectiveness of these methods is still largely constrained by the expressive power of representation learning strategies. }
%In this work, following more recent works~\cite{cea,DBLP:conf/aaai/XuSFSY20}, we focus on the alignment stage instead and strive to improve the decision-making process of alignment to produce more accurate matches. 
%%; that is, assume the features and similarity matrix are fixed, we strive to improve the decision-making process of alignment to produce more accurate matches. 
%More specifically, as shown in Example~\ref{exa:motivate}, the independent decision-making strategy adopted by most \sota approaches neglects the interdependence among the decision-making for each source entity. 
%Intuitively, the information of previously aligned entities would affect the alignment of subsequent entities (e.g., a target entity is less likely to match others if it has already been aligned to a source entity with higher confidence). 
Hence, a few very recent methods~\cite{cea,DBLP:conf/aaai/XuSFSY20} incorporate the 1-to-1 constraint to coordinate the alignment process and align entities collectively. 
This constraint requires that each source entity is matched to exactly one target entity, and vice versa, which has been shown to improve the alignment results. 
However, we find that the decision-making paradigm incorporating the 1-to-1 constraint might not suffice to produce a satisfying alignment, which is shown in Example~\ref{exa:121}. 
%. Let us consider the following example.
This motivates us to find a better characterization of the interdependence. 
Therefore, we propose a new coordination strategy, which is comprised of the \emph{coherence} and \emph{exclusiveness} constraints, to restrict the collective alignment. 
%we propose to two additional constraints---\emph{coherence} and \emph{exclusiveness}---to restrict the collective alignment. %in our RL-based alignment model.
Specifically, coherence aims to keep the EA decisions coherent for closely-related entities. Suppose a source entity $u$ is matched with the target entity $v$; coherence requires that, for the source entities that are closely related to $u$, their corresponding target entities should also be closely related to the target entity $v$. 
Meanwhile, exclusiveness aims to avoid assigning the same target entity to multiple source entities, which requires that, if an entity is already matched, it is less likely to be matched to other entities.
In essence, exclusiveness relaxes the 1-to-1 constraint by allowing some extreme cases, e.g., two source entities both have very high confidence to align the same target entity, and produces a pairing that does not necessarily follow the 1-to-1 constraint.

\begin{example} \label{exa:121}
	Further to Example~\ref{exa:motivate}, by imposing the 1-to-1 constraint,
	%based on the matrix in (b),
	one would arrive at the solution that aligns $u_3$ to $v_2$, since $u_3$ has a higher similarity score with $v_2$ than $u_2$, and hence $u_2$ is compelled to match $v_3$, $u_4$ is compelled to match $v_4$. In this case, although $u_4$ finds the correct match, 
	$u_2$ and $u_3$ still deviate from their correct counterparts.
	%This implies that
	%We find that if we take into account the interconnections between target entities (and source entities), the correct alignment can be recovered.
	
	As for our proposed coordination strategy, the exclusiveness constraint would help prevent $u_2$ from matching $v_1$, since $v_1$ has been confidently aligned to $u_1$; it would also encourage $u_4$ to align $v_4$ rather than $v_2$, since $v_2$ has higher similarity scores with other entities.   
	The coherence constraint would suggest matching $u_2$ with $v_2$, as both $u_2$ and $v_2$ are related with the previously matched pair $(u_1, v_1)$. However, the local similarity score between $(u_3, v_2)$ is much higher than $(u_2, v_2)$.
	In this case, the exclusiveness would allow both $u_2$ and $u_3$ to match entity $v_2$. 
	In comparison with the cases mentioned above, practicing the new coordination strategy reduces the number of incorrect matches. 
	%constraints of coherence and exclusiveness
\end{example} 

To implement the new coordination strategy, we cast the alignment process to the classic sequence decision problem. Given a sequence of source entities, the goal of the sequence decision problem is to decide to which target entity each source entity aligns. 
We approach the problem with a reinforcement learning (RL) based framework, which learns to optimize the decision-making for all entities, rather than optimize every single decision separately. 
More specifically, to implement the coherence and exclusiveness constraints under the RL-based framework, we propose the following design. When making the alignment decision for each source entity, the model takes as input not only local similarity scores, but also coherence and exclusiveness signals generated by previously aligned entities.
Further, we incorporate the coherence and exclusiveness information into the reward shaping.
Thus, the interdependence between EA decisions can be adequately captured.

\myparagraph{Contributions}
This article is an extended version of our previous work~\cite{cea}. In this extension, we make substantial improvement:
\begin{itemize}
	\item  We extend the idea of resolving EA jointly by offering an RL-based collective EA framework (\our) that can model both the \emph{coherence} and \emph{exclusiveness} of EA decisions.
	\item We introduce an adaptive feature fusion strategy to better integrate different features without the need of training data or manual intervention.
	\item We adopt the Bray-Curtis dissimilarity to measure the similarity between entity embeddings, which applies normalization over the calculated distance and leads to better results compared with the commonly used distance measures such as Manhattan distance and cosine similarity.
	\item We add some very recent methods for comparison and conduct a more comprehensive analysis.
\end{itemize}

The main contributions of the article can be summarized as follows:

\begin{itemize}
	\item We identify the deficiency of existing EA methods in making alignment decisions, and propose a novel solution \our to boost the overall EA performance. This is done by
	(1) casting the alignment process into the sequence decision problem, and offering a reinforcement learning (RL) based model to align entities collectively; and
	(2) exploiting representative features to capture different aspects of the similarity between entities, and integrating them with adaptively assigned features.
	
	\item We empirically evaluate our proposal on both cross-lingual and mono-lingual EA tasks against 19 state-of-the-art methods, and the comparative results demonstrate the superiority of \our.
\end{itemize}

\myparagraph{Organization}
In Section~\ref{ref}, we formally define the task of EA and introduce related work. 
In Section~\ref{method}, we present the outline of \our. 
In Section~\ref{fg}, we introduce the feature generation process. 
In Section~\ref{ff}, we introduce the adaptive feature fusion strategy. 
In Section~\ref{collective}, we elaborate the RL-based collective EA strategy. 
In Section~\ref{exp}, we introduce the experimental settings. 
In Section~\ref{rna}, we report and analyze the results. 
In Section~\ref{conclude}, we present the conclusion and future works.

\section{Preliminaries}
\label{ref}
In this section, we first formally define the task of EA, and then introduce the related work.
%the possibly relevant tasks and their differences from EA. Finally, we briefly present state-of-the-art EA solutions.

\subsection{Task Definition}
The task of EA aims to align entities in different KGs. 
A KG $G = (E, R, T)$ is a directed graph comprising a set of entities $E$, relations $R$, and triples $T$. A triple $t = (e_i, r_{ij}, e_j) \in T$ represents that a head entity $e_i$ is connected to a tail entity $e_j$ via a relation $r_{ij}$. 

The inputs to EA are a source KG $G_1 = (E_1, R_1, T_1)$, a target KG $G_2 = (E_2, R_2, T_2)$, a set of seed entity pairs $S = \{(u_s,v_s)|u_s\in E_1^s, v_s\in E_2^s, u_s \leftrightarrow v_s\}$, where $E_1^s$ and $E_2^s$ denote the source and target entities in the training set, respectively, $u_s \leftrightarrow v_s$ indicates the source entity $u_s$ and the target entity $v_s$ are \emph{equivalent}, i.e., $u_s$ and $v_s$ refer to the same real-world object. 
The task of EA is defined as finding a target entity $v^*$ for each source entity $u$ in the test set, i.e., $\Psi = \{(u,v^*)|u\in E_1^t, v^*\in E_2^t\}$, where $E_1^t$ and $E_2^t$ represent the source and target entities in the test set, respectively. 
A pair of aligned entities $\phi\in\Psi$ is also called a \emph{correspondence}. 
If the source and target entities in the \emph{correspondence} are \emph{equivalent}, we call it a \emph{correct correspondence}.

\subsection{Related Work}
\myparagraph{Entity Resolution and Instance Matching}
While the problem of EA was introduced a few years ago, the more generic version of the problem---identifying entity records referring to the same real-world entity from different data sources---has been investigated from various angles by different communities~\cite{9174835}.
%The general problem of identifying entity records referring to the same real-world entity from different data sources has been investigated from various angles.
It is mainly referred to as entity resolution
(ER)~\cite{DBLP:conf/cikm/NieHHSCZWK19,DBLP:conf/ijcai/FuHSCZWK19}, entity
matching~\cite{DBLP:conf/sigmod/DasCDNKDARP17,DBLP:conf/sigmod/MudgalLRDPKDAR18} or record linkage~\cite{DBLP:journals/tkde/Christen12}.
These tasks assume the inputs are \emph{relational data}, and each data object usually has a large amount of textual information described in multiple attributes. 
Since EA pursues the same goal as ER, it can be deemed a special but non-trivial case of ER, which aims to handle KGs and deal exclusively with binary relationships, i.e., graph-shaped data~\cite{9174835}.

We are aware that there are some collective ER approaches targeted at graph-structured data~\cite{DBLP:journals/tkdd/BhattacharyaG07}, represented by \paris~\cite{DBLP:journals/pvldb/SuchanekAS11} and \textsf{SiGMa}~\cite{DBLP:conf/kdd/Lacoste-JulienPDKGG13}. 
To model the relations among entity records, they adopt collective alignment algorithms such as similarity propagation~\cite{DBLP:journals/pvldb/AltowimKM14}, the LDA model~\cite{DBLP:conf/sdm/BhattacharyaG06}, the conditional random fields model~\cite{DBLP:conf/nips/McCallumW04}, Markov logic network models~\cite{DBLP:conf/icdm/SinglaD06}, or probabilistic soft logic~\cite{DBLP:journals/kais/KoukiPMKG19}. 
These approaches are frequently related to instance matching (or A-Box matching), which aims to find correspondences between entities in different ontologies.
%, a key issue in the process of integrating heterogeneous data sources described by ontologies.
Since these methods are established in a setting similar to EA, we include them in the experimental study, and use \emph{collective ER approaches} as the general reference to them. 
Note that different from these methods, EA solutions build on the recent advances in deep learning and mainly rely on graph representation learning technologies to model the KG structure and generate entity embeddings for alignment~\cite{9174835}.

\myparagraph{Entity Alignment}
A shared pattern can be observed from current EA approaches. 
First, they generate for each feature a unified embedding space where the entities from different KGs are directly comparable. 
%The first step is to generate feature-specific entity embeddings. 
A frequently used feature is the KG structure, as the equivalent entities in different KGs tend to possess very similar neighboring information. 
To generate structural entity embeddings, \emph{structure encoders} are devised, including KG representation based models~\cite{DBLP:conf/ijcai/ChenTYZ17,DBLP:conf/ijcai/ZhuXLS17,DBLP:conf/ijcai/SunHZQ18,DBLP:conf/ijcai/ZhuZ0TG19,DBLP:conf/semweb/SunHL17,DBLP:conf/aaai/TrisedyaQZ19,DBLP:conf/ijcai/ChenTCSZ18}, e.g., \transe~\cite{DBLP:conf/nips/BordesUGWY13}, and graph neural network (GNN) based models~\cite{DBLP:conf/acl/CaoLLLLC19,DBLP:conf/emnlp/LiCHSLC19,DBLP:conf/emnlp/WangLLZ18,ACL19,DBLP:conf/ijcai/WuLF0Y019}, e.g., GCN~\cite{DBLP:journals/corr/KipfW16}. 
Other available features include attributes~\cite{DBLP:conf/semweb/SunHL17,DBLP:conf/emnlp/WangLLZ18,DBLP:conf/aaai/TrisedyaQZ19,DBLP:conf/emnlp/YangZSLLS19,IJCAI19}, entity names~\cite{ACL19,DBLP:conf/ijcai/WuLF0Y019,IJCAI19} and entity descriptions~\cite{DBLP:conf/ijcai/ChenTCSZ18,DBLP:conf/emnlp/YangZSLLS19}. 
Correspondingly, \emph{additional information encoders} are designed to embed these features into vector representations. % for entities.  
To project the feature-specific embeddings from different KGs into a unified space, they devise unification functions, e.g., the margin-based loss function~\cite{DBLP:conf/acl/CaoLLLLC19,DBLP:conf/emnlp/LiCHSLC19,DBLP:conf/emnlp/WangLLZ18,DBLP:conf/emnlp/YangZSLLS19,DBLP:conf/ijcai/WuLF0Y019,DBLP:conf/emnlp/WuLFWZ19} and the transition function~\cite{DBLP:conf/ijcai/ChenTYZ17,DBLP:conf/ijcai/ZhuXLS17,DBLP:conf/ijcai/ChenTCSZ18}, or exploit the corpus fusion strategy~\cite{DBLP:conf/icml/GuoSH19,DBLP:conf/ijcai/SunHZQ18,DBLP:conf/ijcai/ZhuZ0TG19}. 

Then, they determine the most likely target entity for each source entity, given the feature-specific unified embeddings. 
The most common approach is to return a ranked list of target entities for each source entity according to a specific \emph{distance measure} between embeddings, among which the top ranked entity is regarded as the match. 
Frequently used distance measures include the Euclidean distance~\cite{DBLP:conf/ijcai/ChenTYZ17,DBLP:conf/ijcai/ZhuXLS17,DBLP:conf/emnlp/YangZSLLS19}, the Manhattan distance~\cite{DBLP:conf/ijcai/WuLF0Y019,DBLP:conf/emnlp/WuLFWZ19,DBLP:conf/emnlp/WangLLZ18} and the cosine similarity~\cite{DBLP:conf/semweb/SunHHCGQ19,DBLP:conf/semweb/SunHL17,DBLP:conf/ijcai/SunHZQ18}~\footnote{Note that the distance score between entities can be easily converted to the similarity score by subtracting the distance score from 1. 
	Therefore, in this paper, we may use \emph{distance} and \emph{similarity} interchangeably.}. 
If multiple features are adopted, it is of necessity to aggregate these features. 
Some \emph{feature fusion} methods to this end are \emph{representation-level}, which directly combine the feature-specific embeddings and generate an aggregated entity embedding for each entity~\cite{ACL19,DBLP:conf/ijcai/WuLF0Y019,DBLP:conf/emnlp/WuLFWZ19,IJCAI19}. Then the similarity between the aggregated entity embeddings is used to characterize the similarity between entities.
Other approaches are \emph{outcome-level}, which combine the feature-specific similarity scores with hand-tuned weights~\cite{DBLP:conf/emnlp/WangLLZ18,DBLP:conf/esws/PangZTT019}. 
%A few studies~\cite{cea,DBLP:conf/aaai/XuSFSY20} notice that separately generating alignment result for each source entity neglects the interdependence between EA decisions. 
%%might introduce false many-to-1 matches. 
%Hence, they employ the 1-to-1 constraint to make \textbf{collective alignment}. 
Notably, some very recent works~\cite{cea,DBLP:conf/aaai/XuSFSY20} propose to exert the 1-to-1 constraint into the alignment process, which can align entities jointly and generate more accurate results. Specifically, \cea~\cite{cea} formulates EA as a stable matching problem~\cite{gale1962college}, and uses the deferred acceptance algorithm~\cite{DBLP:journals/ijgt/Roth08} to produce the results, i.e., no pair of two entities from the opposite side would prefer to be matched to each other rather than their assigned partners~\cite{DBLP:conf/ccs/DoernerES16}. 
\gmehd~\cite{DBLP:conf/aaai/XuSFSY20} views EA as a task assignment problem, and employs the Hungarian algorithm~\cite{kuhn1955hungarian} to find a set of 1-to-1 alignments that maximize the total pair-wise similarity.

There are some recent studies that improve EA results with the bootstrapping strategy~\cite{DBLP:conf/ijcai/SunHZQ18,DBLP:conf/ijcai/ZhuXLS17} or relation modeling~\cite{DBLP:conf/emnlp/WuLFWZ19,DBLP:conf/wsdm/MaoWXLW20,DBLP:conf/aaai/SunW0CDZQ20}. 

\myparagraph{Reinforcement Learning}
Over recent years, reinforcement learning (RL) has been frequently used in many sequence decision problems~\cite{DBLP:conf/www/FangC0ZZL19}. 
Clark and Manning apply RL on coreference resolution to directly optimize a neural mention-ranking model for coreference evaluation metrics, which avoids the need for carefully tuned hyper-parameters~\cite{DBLP:conf/emnlp/ClarkM16}. 
Fang et al. convert entity linking into a sequence decision problem and use an RL model to make decisions from a global prospective~\cite{DBLP:conf/www/FangC0ZZL19}. 
Feng et al. propose an RL-based model for sentence-level relation classification from noisy data, where a selection decision is made for each sentence in the sentence sequence using a policy network~\cite{DBLP:conf/aaai/FengHZYZ18}.
Inspired by these efforts, we also consider EA as the sequence decision process and harness RL to make alignment decisions.

\section{The Framework of Our Proposed Model}
\label{method}
\begin{figure*}[h]
	\centering
	\includegraphics[width=0.75\linewidth]{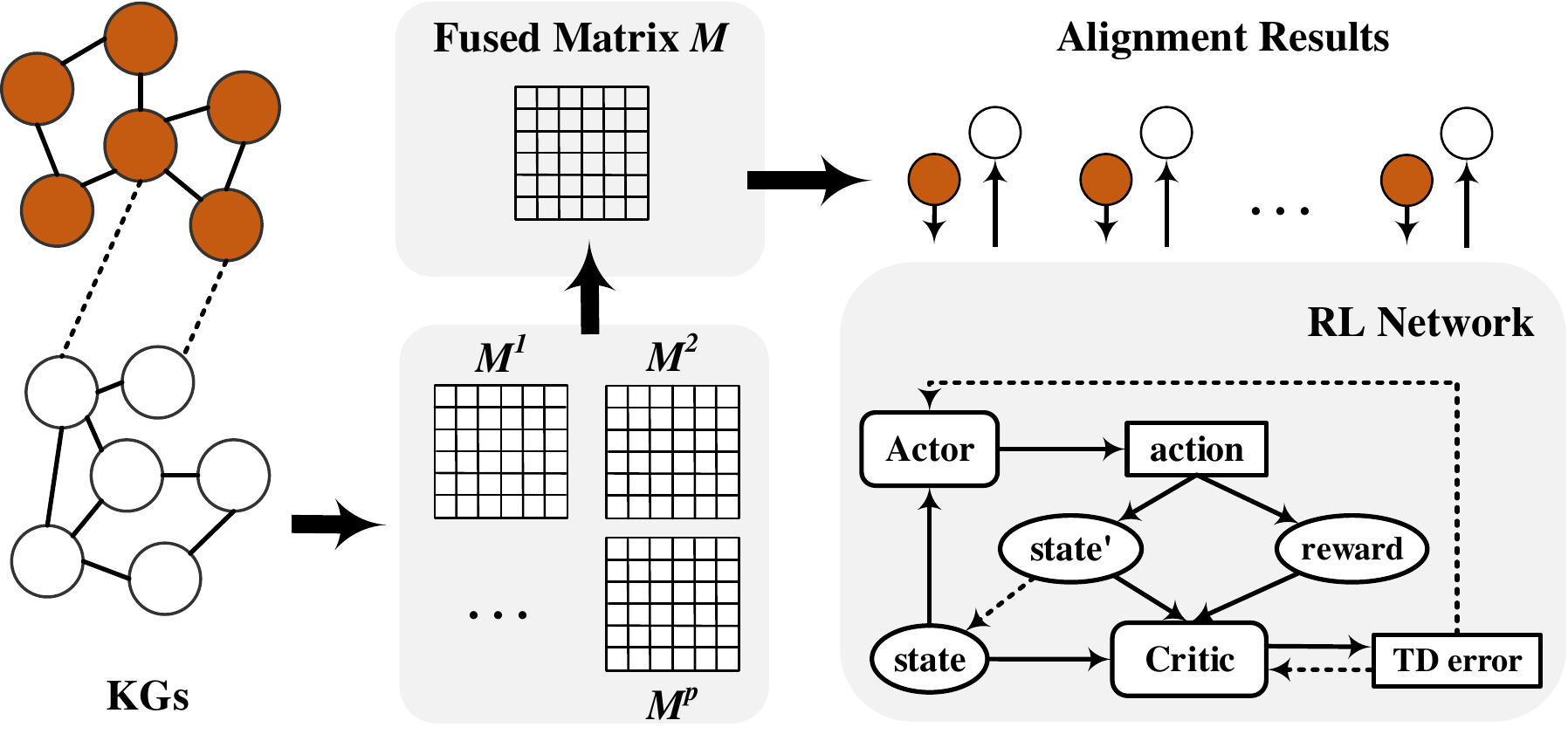}
	\caption{The framework of \our. 
		The dashed lines in the RL Network represent the \emph{update} operation.
	}
	\label{fig:framework}%
\end{figure*}

%\myparagraph{Outline of Proposed Framework}
%We then introduce the outline of our proposal, \our.
%, which aims to collectively align entities based on adaptively fused features. 
As shown in Figure~\ref{fig:framework}, there are three stages in our proposed framework: 

\begin{itemize}
	\item \textit{Feature Generation}. This stage generates representative features for EA, including the structural embeddings learned by GCN, the semantic embeddings represented as averaged word embeddings, and the string similarity matrix calculated using the Levenshtein distance~\cite{levenshtein1966binary}. 
	%	This component corresponds to the \emph{Pre-alignment Module}, which is not the focus of this study. 
	
	\item \textit{Adaptive Feature Fusion}. 
	This stage fuses various features with adaptively generated weights. 
	We first convert the structural and semantic representations into corresponding similarity matrices using the Bray-Curtis dissimilarity. 
	Then, we combine different features with adaptively assigned weights and generate a fused similarity matrix. 
	%	This step corresponds to the \emph{distance measure} and \emph{feature fusion} components in the \emph{Alignment Module}.
	%	, where we make novel contributions.
	
	\item \textit{Collective EA}. 
	This stage takes the fused similarity matrix as input and generates alignment results by capturing the interdependence between decisions for different entities. 
	Concretely, we convert EA into the sequence decision problem and adopt a deep RL network to model both the coherence and exclusiveness of EA decisions. 
	%	This step corresponds to the \emph{collective alignment} part in the \emph{Alignment Module}.
	%	, which has been rarely explored.
\end{itemize}

Next, we elaborate the modules of \our. 

\section{Feature Generation}
\label{fg}
In this section, we introduce the features employed in \our.

\subsection{Structural Information}
\label{stru}
We harness the graph convolutional network (GCN)~\cite{DBLP:journals/corr/KipfW16} to encode the neighborhood information of entities as real-valued vectors.\footnote{We reckon that there are more advanced frameworks, e.g., the dual-graph convolutional network~\cite{DBLP:conf/ijcai/WuLF0Y019} and the recurrent skipping network~\cite{DBLP:conf/icml/GuoSH19}, for learning the structural representation. 
They can also be easily plugged into our overall framework after parameter tuning. 
We will explore these options in the future, as they are not the focus of this article.}   
Then, we briefly introduce the configuration of GCN for the EA task, and leave out the GCN fundamentals in the interest of space. 

\myparagraph{Model Configuration and Training}
GCN is harnessed to generate structural representations of entities. 
We build two 2-layer GCNs, and each GCN processes one KG to generate the embeddings of entities. 
The initial feature matrix, $\vec X$, is sampled from the truncated normal distribution with L2-normalization on rows.\footnote{Other methods of initialization are also viable. We stick to the random initialization in order to capture the ``pure'' structural signal.} 
It gets updated by the GCN layers, and thus the final output matrix $\vec Z$ can encode the neighboring information of entities.
%encode the structural information contained in each KG. 
The adjacency matrix $\vec A$ is constructed according to~\cite{DBLP:conf/emnlp/WangLLZ18}.
Note that the dimensionality of the feature vectors is fixed at $d_s$ and kept the same for all layers. 
The two GCNs share the same weight matrix in each layer. 
%, $\vec W^1$ and $\vec W^2$,

Then, we project the entity embeddings generated by two GCNs into a unified embedding space using pre-aligned EA pairs $S$. 
Specifically, the training objective is to minimize the margin-based ranking loss function:
\begin{equation}
\label{eq:1}
L = \sum_{(u_s,v_s)\in S} \sum_{(u^{\prime}_s,v^{\prime}_s)\in S_{(u_s,v_s)}^{\prime}} [  \parallel \vec u_s-\vec v_s\parallel_{l1} - \parallel\vec{u_s^{\prime}}-\vec {v_s^{\prime}}\parallel_{l1} + \epsilon]_+ ,
\end{equation}
where $[x]_+ = max\{0,x\}$, $S_{(u_s,v_s)}^{\prime}$ denotes the set of negative EA pairs obtained by corrupting $(u_s,v_s)$, i.e., substituting $u_s$ or $v_s$ with a randomly sampled entity from its corresponding KG. 
$\vec u_s$ and $\vec v_s$ denotes the (structural) embedding of entity $u_s$ and $v_s$, respectively. 
$\epsilon$ is a positive margin that separates the positive and negative EA pairs. Stochastic gradient descent is harnessed to minimize the loss function.

\subsection{Semantic Information}
\label{stri}
% tends to be
For each entity, there is textual information associated with it, ranging from its name, its description, to its attribute values. % in the textual form. 
This information can help match entities in different KGs, since equivalent entities share the same meaning. 
Among this information, the entity name, which identifies an entity, is the most universal textual form. 
Also, given two entities, comparing their names is the easiest approach to judge whether they are the same. 
Therefore, in this work, we propose to utilize entity names as the source of textual information.
% for EA.

Entity names can be exploited from the semantic- and string-level.  
We first introduce the \emph{semantic similarity}. 
%, as it can also work when the vocabularies of KGs differ, especially for the cross-lingual scenario. 
More specifically, we use the averaged word embeddings to capture the semantic meaning of entity names.
For a KG, the name embeddings of all entities are denoted in matrix form as $\vec N$. 
Like word embeddings, similar entity names will be placed adjacently in the entity name representation space. 

\subsection{String Information}
Current methods mainly capture the semantic information of entities. 
In this work, we contend that, the string information, which has been largely overlooked by current EA literature, is also a contributive feature, since: 
\begin{inparaenum} [(1)]
	\item string similarity is especially useful in tackling the mono-lingual EA task or the cross-lingual EA task where the languages of the KG pair are closely-related (e.g., English and German);  and 
	\item string similarity does not rely on external resources, e.g., pre-trained word embeddings. 
\end{inparaenum}
In particular, we adopt the Levenshtein distance~\cite{levenshtein1966binary}, a string metric for measuring the difference between two sequences. 
We denote the string similarity matrix calculated by the Levenshtein ratio as $\vec {M^l}$. 

\section{Adaptive Feature Fusion}
\label{ff}
In this section, we first introduce a new measure to capture the similarity between entity embeddings. Then we point out the limitations of current fusion methods. 
Finally, we elaborate our proposed adaptive feature fusion strategy.

\subsection{Distance Measures}
\label{Dis}
Existing solutions characterize the similarity between entity embeddings by the Manhattan distance~\cite{DBLP:conf/ijcai/WuLF0Y019,DBLP:conf/emnlp/WuLFWZ19,DBLP:conf/emnlp/WangLLZ18}, the Euclidean distance~\cite{DBLP:conf/ijcai/ChenTYZ17,DBLP:conf/ijcai/ZhuXLS17,DBLP:conf/emnlp/YangZSLLS19},  and the cosine similarity~\cite{DBLP:conf/semweb/SunHHCGQ19,DBLP:conf/semweb/SunHL17,DBLP:conf/ijcai/SunHZQ18}. 
Given two entities $u$ and $v$ with embeddings $\vec u=(u_1, u_2,..., u_n)$ and $\vec v=(v_1, v_2,..., v_n)$, the Manhattan distance is formalized as:
\begin{equation}
D_m(u,v) = \sum_{i=1}^{n}\left|u_i - v_i\right|,
\end{equation}
and the corresponding similarity score is $1- D_m(u,v)$. Similarly, the Euclidean distance is:
\begin{equation}
D_e(u,v) = \sum_{i=1}^{n} \left\|u_i - v_i\right\|_2,
\end{equation}
and the corresponding similarity score is $1- D_e(u,v)$.
Nevertheless, these measures fail to apply normalization over the calculated distance. 
As a remedy, we use the Bray-Curtis dissimilarity %~\cite{bray1957ordination} 
to measure the distance between entity embeddings: 
%The normalization is done using the absolute difference divided by the summation:
\begin{equation}
D_b(u,v) = \sum_{i=1}^{n} \frac{\left|u_i - v_i\right|}{\left|u_i + v_i\right|},
\end{equation}
and accordingly, the similarity score is $1- D_b(u,v)$. 
%Bray-Curtis dissimilarity is not a metric as it does not satisfy the triangle inequality property, whereas it brings better empirical results by performing normalization over the absolute distance, as shown in Section~\ref{exp}. 

%Admittedly, the cosine similarity also applies normalization to obtain the similarity score between two embeddings:
Admittedly, the cosine similarity also applies normalization to obtain the similarity score between two embeddings:
\begin{equation}
Sim_c(u,v) = \frac{\vec u\cdot\vec v}{\left\|\vec u\right\|_2\left\|\vec v\right\|_2}.
\end{equation}

However, unlike the Bray-Curtis dissimilarity that performs normalization for \emph{each pair} of elements in the vectors, the cosine similarity directly normalizes the dot product of two vectors. 
We empirically evaluate these distance measures, and demonstrate the superiority of the Bray-Curtis dissimilarity in Section~\ref{rna:rq4}. 

Given the learned structural embedding matrix $\vec Z$ and the entity name embedding matrix $\vec N$ from Section~\ref{stru} and~\ref{stri}, respectively, we use the Bray-Curtis dissimilarity to generate pair-wise similarity scores between entities. 
We denote the resultant structural similarity matrix as $\vec {M^s}$, where rows represent the source entities in the test set, columns denote the target entities in the test set, and each element in the matrix denotes the structural similarity score between a pair of source and target entities. Similarly, the resultant semantic similarity matrix is represented as $\vec {M^n}$.

\subsection{Feature Fusion}
\label{sect:ff}
\begin{figure}
	\centering
	\includegraphics[width=0.9\linewidth]{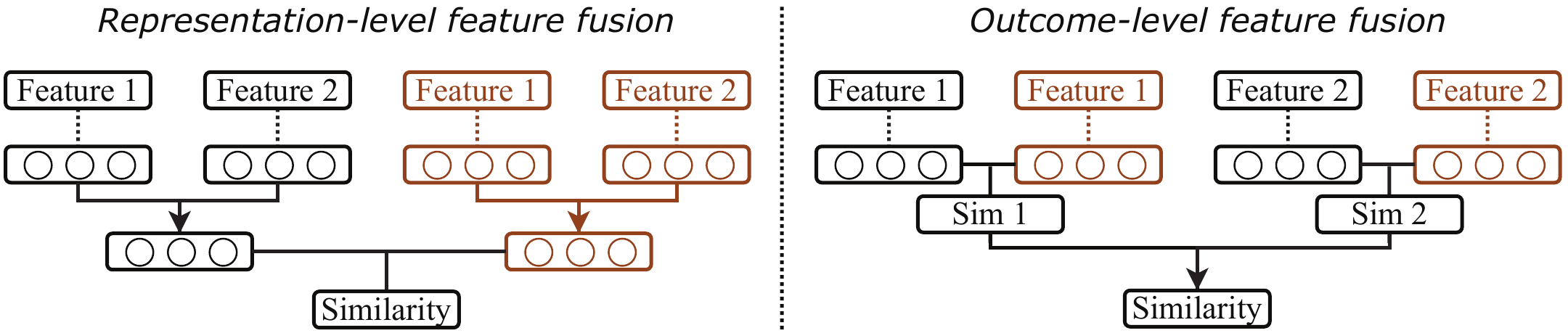}
	\caption{Different strategies of feature fusion. The operations for different entities are represented in different colors. Dashed lines denote generating vector representations of features. Solid lines without arrows represent calculating the similarity between embeddings, while solid lines with arrows represent the feature fusion process.}
	\label{fig:fu}
\end{figure}

\myparagraph{Different Strategies of Feature Fusion}
To fuse different features, the state of the art directly combines feature-specific embeddings and generates an aggregated embedding for each entity, which is then used to calculate pair-wise similarity between entities (shown in the left of Figure~\ref{fig:fu}). 
Specifically, some design aggregation strategies to integrate multiple view-specific entity embeddings~\cite{IJCAI19}, while some treat certain features as inputs for learning representations of other features~\cite{ACL19,DBLP:conf/ijcai/WuLF0Y019}. 
Nevertheless, these \emph{representation-level} fusion techniques might fail to maintain the characteristics of the original features, e.g., two entities might be extremely similar in feature-specific embedding spaces, while placed distantly in the unified representation space. 

In this work, we resort to the \emph{outcome-level} feature fusion, which operates on the intermediate outcomes---feature-specific similarity matrices. 
Current approaches to this end first calculate pair-wise similarity scores within each feature-specific space, and then combine these scores to generate the final similarity score~\cite{DBLP:conf/emnlp/WangLLZ18,DBLP:conf/semweb/SunHL17,cea} (shown in the right of Figure~\ref{fig:fu}).
However, they manually assign the weights of features, 
%in these \emph{outcome-level} feature fusion strategies, the weights of features are manually assigned, 
which can be inapplicable when the number of features increases, or the importance of certain features varies greatly under different settings. 
Therefore, it is of significance to dynamically determine the weight of each feature.  

\myparagraph{Adaptive Feature Fusion Strategy}
In this work, we offer an adaptive feature fusion strategy that can dynamically determine the weights of features without the training data. It consists of the following stages:

\begin{figure}[h]
	\centering
	\includegraphics[width=0.8\linewidth]{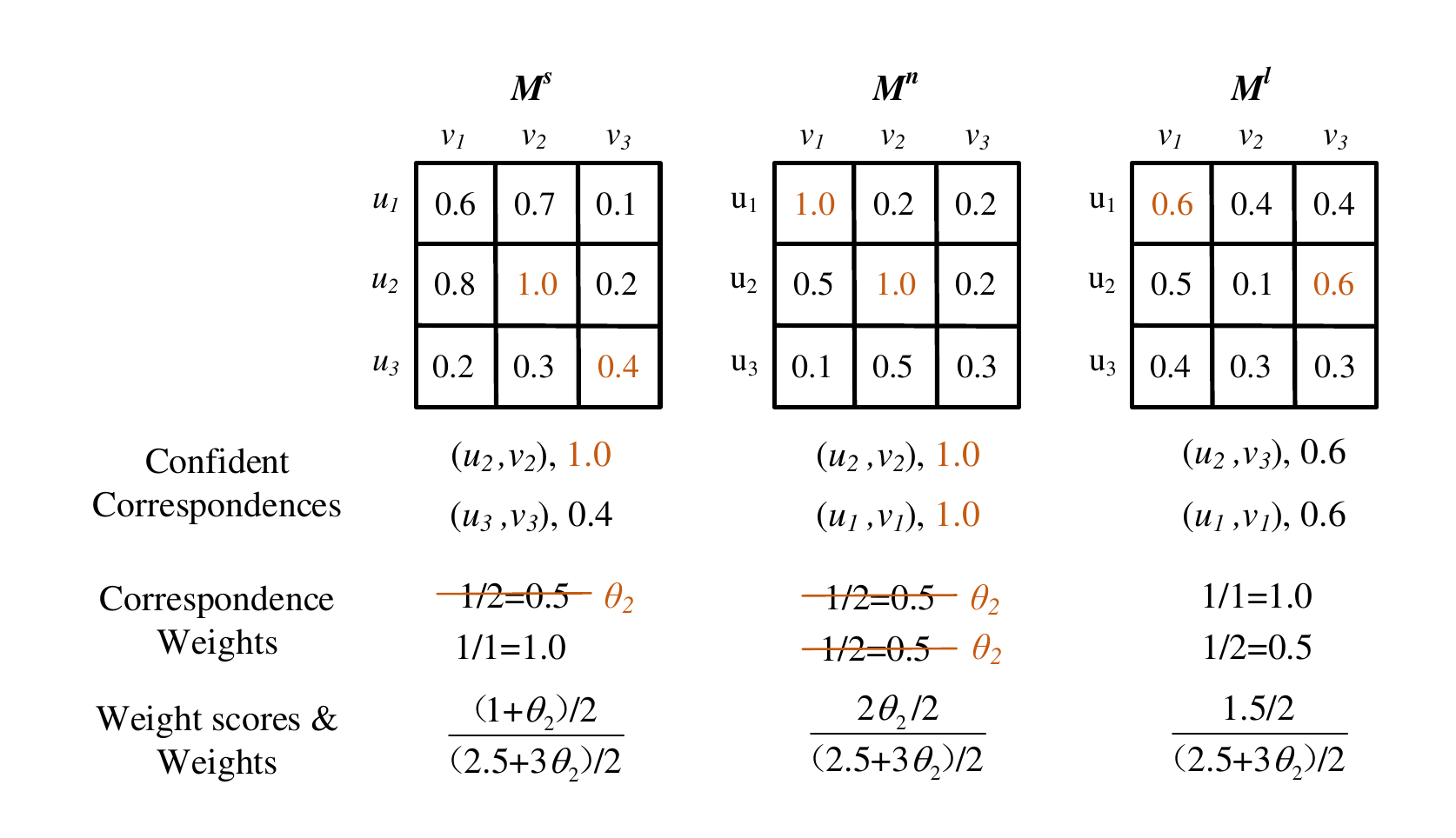}
	\caption{The framework of adaptive weight assignment.}
	\label{fig:agg}%
\end{figure}

\subsubsection*{Confident Correspondence Generation}
The inputs to this stage are $p$ features and their corresponding similarity matrices---$\vec {M^1}, \vec {M^2}, ..., \vec {M^p}$. 
$\vec M^p_{ij}$ denotes the similarity between the source entity $u_i$ and the target entity $v_j$, measured by the feature $p$. 
If $\vec M^p_{ij}$ is the largest both along the row and the column, we consider $(u_i, v_j)$ to be a \emph{confident correspondence} generated by the feature $p$. 
This is a relatively strong constraint, and the resulting correspondences are very likely to be correct. 
As thus, we assume that the confident correspondences generated by a specific feature can reflect its importance. The concrete correlation is formulated later. 

As shown in Figure~\ref{fig:agg}, there are three feature matrices, $\vec {M^s}, \vec {M^n}, \vec {M^l}$, each of which generates two confident correspondences with similarity scores.

\subsubsection*{Correspondence Weight Calculation}
Then, we determine the weight of each confident correspondence. 
Instead of assigning equal weights, we assume that the importance of a correspondence is inversely proportional to the number of its occurrences.
Specifically, if a correspondence is generated by $q$ features, its weight is set to $1/q$, as it is believed that the frequently occurring correspondence brings less new information in comparison with a correspondence that has the quality of being detected by only one single feature matrix~\cite{DBLP:journals/ws/GulicVB16}. 

Additionally, for a correspondence with very large similarity score $\vec M^p_{ij} > \theta_1$, we set its weight to a small value $\theta_2$. 
%This is to prevent a very unbalanced distribution of weights when a specific feature is extremely effective. 
With this setting, the features that are very effective would not be assigned with extremely large weights, and the less effective features can still contribute to the alignment. 
We empirically validate the usefulness of this strategy in Section~\ref{rna:rq3}.
%will further discuss the motivation and the setting of $\theta_1, \theta_2$ in 

As shown in Figure~\ref{fig:agg}, regarding the two correspondences generated by $\vec {M^s}$, $(u_3, v_3)$ is assigned with weight $1/1 = 1$, as it is only produced by $\vec {M^s}$, while $(u_2, v_2)$ is detected by both $\vec {M^s}$ and $\vec {M^n}$, and hence endowed with the weight $1/2 = 0.5$. Moreover, the weight of $(u_2, v_2)$ is reset to $\theta_2$, since its similarity score exceeds $\theta_1 = 0.95$.

%\myparagraph{Feature Weight Calculation}
\subsubsection*{Feature Weight Calculation}
After obtaining the weight of each confident correspondence, the \emph{weight score} of feature $p$ is defined as the sum of the weights of its confident correspondences, divided by the number of these correspondences. 
The \emph{weight} of feature $p$ is the ratio between its weight score and the total weight score of all features.

%\myparagraph{Feature Fusion with Adaptive Weight}
\subsubsection*{Feature Fusion with Adaptive Weight}
We combine the feature-specific similarity matrices with their adaptively assigned weights to generate the fused similarity matrix $\vec {M}$. 
Particularly, in this work, we utilize three representative features---structural-, semantic- and string-level features, which are elaborated in Section~\ref{fg}. 

\myparagraph{Remark}
One possible solution to determine the weight of each feature is using machine learning techniques to learn the weights. 
Nevertheless, for each source entity, the number of negative target entities is much larger than the positive ones. 
Besides, the amount of training data is very limited. 
Such restraints make it difficult to generate high-quality training corpus, and hence might hamper learning appropriate weights. 
We will report the performance of the learning-based approach in Section~\ref{rna:rq3}.

\section{RL-based Collective EA Strategy}
\label{collective}
After obtaining the fused similarity matrix $\vec M$, we proceed to the alignment decision-making stage, which is the core step of the EA task and can be regarded as a matching process, i.e., matching a source entity with a target entity. 
The majority of existing works choose the optimal local match for each source entity while neglecting the influence between matching decisions. 
To capture the interdependence among alignment decisions, \cea~\cite{cea} and \gmehd~\cite{DBLP:conf/aaai/XuSFSY20} exert the 1-to-1 constraint on the matching process. 
More specifically, \cea adopts the deferred acceptance algorithm to find a stable matching result, which has the worst runtime complexity of $O(n^2)$. 
The resulting matching is locally optimal, i.e., a man-optimal assignment (the man refers to the source entity in our case). 
Meanwhile, \gmehd frames the matching process as the task assignment problem (maximizing the local similarity scores under the 1-to-1 constraint), which is essentially a fundamental combinatorial optimization problem whose exact solution can be found by the Hungarian algorithm. 
The Hungarian algorithm has the worst runtime complexity of $O(n^3)$, and \gmehd employs a search space separation strategy to reduce the time cost (which also hurts the performance). 

There are several main drawbacks of framing the alignment process as the matching problem: (1) High worst time complexity: $O(n^2)$ for the deferred acceptance algorithm and $O(n^3)$ for the Hungarian algorithm; and 
(2) Difficulty of including other constraints. As stated in the introduction, the coherence constraint is also of importance. Nevertheless, it is hard to incorporate such constraint into the stable matching process. For the combinatorial optimization problem, although we can consider the coherence and define the goal as maximizing the local similarity scores \emph{and} the overall coherence, the resulting optimization problem becomes NP-hard (similar to the global optimization in Entity Linking task~\cite{DBLP:journals/tkde/ShenWH15}). 
Besides, there is no straightforward approach to replace the 1-to-1 constraint in the matching process with the exclusiveness constraint. 

%we can determine the EA results in an independent fashion, which has been adopted by \textcolor{blue}{most state-of-the-art methods~\cite{DBLP:conf/aaai/SunW0CDZQ20,DBLP:conf/wsdm/MaoWXLW20,9174835}.} 
%Concretely, for each source entity $u$, we retrieve its corresponding row entry in $\vec M$, and rank the elements in a descending order. 
%The top-ranked target entity is aligned to the source entity. 
%Nevertheless, this approach fails to consider the interdependence between different EA decisions. 
%Although some efforts~\cite{cea,DBLP:conf/aaai/XuSFSY20} partially address this issue by adding the 1-to-1 constraint into the alignment process, they still cannot adequately model the dynamics. % between different EA decisions. 
%To fill in this gap, we offer an RL-based collective alignment strategy. 

\subsection{EA as a Sequence Decision Problem}
%As a result, we cast the alignment process to the sequential decision problem. Since there are no supervisory signals, we use RL to find the set of actions (alignment results) and incorporate several constraints into the reward shaping to provide more accurate guidance. 
%The experimental results and feature analysis validate that using RL to tackle EA can achieve superior empirical alignment results than using the matching algorithms. 
%Note that there are also a few works that employ RL to tackle graph matching and combinatorial optimization problems and improve the state-of-the-art performance~\cite{DBLP:conf/nips/KhalilDZDS17,DBLP:conf/icde/WangTLXXL19}.
In this work, we cast EA to the classic sequence decision problem; that is, given a sequence of source entities, EA generates for each source entity a target entity. 
This problem can be solved by a reinforcement learning (RL) based model.
In this model, source entities are aligned in a sequential but collective manner, which allows every decision to be made based on current state and previous ones, such that the interdependence among the decisions is captured.
Particularly, we characterize the interdependence by a novel coordination strategy, which comprises  
%The interdependence is characterized by 
%To characterize the interdependence, we devised a new coordination strategy, which comprises 
%To accurately generate the results, we use an RL network to model the dynamics during the alignment process, i.e., 
the \emph{exclusiveness} constraint---a previously chosen target entity is less likely to be assigned to the rest of the source entities, and the \emph{coherence} constraint---the relevance between entities can assist the decision-making.

%More specifically, 
%Regarding the choice of the RL model, 
As for the RL framework, 
we adopt the Advantage Actor-Critic (A2C)~\cite{DBLP:conf/icml/MnihBMGLHSK16} model, where the Actor conducts actions in an environment and the Critic computes the value functions to help the actor in learning. 
These two agents participate in a game where they both get better in their own roles as the learning gets further. 
As shown in Figure~\ref{fig:framework}, at each step, the RL network takes the current state as input, outputs an action (a target entity), which influences the following state and generates a reward. Taking the current state, the next state, and the reward as input, the Critic calculates the Temporal-Difference (TD) error and updates its network. The TD error is then fed to the Actor for the parameter updates. 
%This procedure continues until the framework reaches the final state.
%The result is that the overall architecture will learn to play the game more efficiently than the two models separately. 

Next, we introduce the basic components of the RL model, as well as the optimization and learning process.
%, including the environment, state, action, and reward.
%~\cite{DBLP:conf/nips/KondaT99}

\subsection{RL-based Collective EA Model}

\myparagraph{Environment}
The environment includes the source and target entities in the test set, the adjacency matrices of KGs, as well as the fused similarity matrix $\vec M$. 

\myparagraph{State}
The state vector $\vec s$ is expressed as $\vec{s^1} \circ \vec{s^2} + \vec{s^3}$, where $\circ$ represents element-wise product, and $\vec{s^1}$, $\vec{s^2}$, $\vec{s^3}$ refer to the \emph{local similarity vector}, the \emph{exclusiveness vector} and the \emph{coherence vector}, respectively. More specifically, 
\begin{itemize}
	\item The \emph{local similarity vector} is obtained from the fused similarity matrix $\vec M$, which characterizes the similarity between the current source entity and the candidate target entities.
	%	 in terms of the available features for alignment. 
	
	\item The \emph{exclusiveness vector} indicates the target entities that have been chosen. 
	Each element in the vector corresponds to a target entity, and the value can be chosen between ``1'' and ``-1''. While ``1'' denotes that the corresponding target entity has not been chosen yet, ``-1'' denotes that the corresponding target entity has been chosen. 
	The exclusiveness vector is initialized with ``1''s; then if a target entity is chosen, the value in the corresponding position is replaced with ``-1''. 
	
	\item The \emph{coherence vector} characterizes the relevance between current candidate target entities and previously chosen target entities. 
	Concretely, given the current source entity, we first retrieve its related source entities that have been matched according to the adjacency matrix in the source KG.
	We consider the target entities chosen by these source entities as the \emph{contextual entities} for aligning the current source entity. 
	For each candidate target entity, we define its coherence score as the number of \emph{contextual entities} it is directly connected with in the KG. %connects with. 
	The coherence scores of all candidate target entities constitute the \emph{coherence vector}. 
\end{itemize}

\myparagraph{Action}
An action denotes the Actor choosing a target entity for a source entity given a particular state. 
The Actor takes the current state as input, and chooses a target entity from the candidate target entities according to the probabilities calculated by a neural model.
The neural network (parameterized by $\theta$) consists of a fully connected layer and a softmax layer. 
More specifically, 
%The structure of the neural network is shown in Figure. 
the input state at step $i$ is fed into the fully connected layer, which generates a hidden state:
\begin{equation}
\label{eq:actor}
\vec{h}(s_i) = Relu(\vec{W_1}\vec{s_i} + \vec{b_1}),
\end{equation}
where $\vec{W_1}$ and $\vec{b_1}$ are the parameters of this layer. The hidden state is then fed to the softmax layer, which generates the probability distribution of actions:
\begin{equation}
\pi_\theta(a|s_i) = Softmax(\vec{W_2}\vec{h}({s_i}) + \vec{b_2}),
\end{equation}
where $\vec{W_2}$ and $\vec{b_2}$ are the parameters of the softmax layer.
Then, the target entity is sampled from the candidate target entities, according to $\pi_\theta(a|s_i)$.
Notably, in order to reduce the model complexity and increase the efficiency, for each source entity, we merely consider the top-$\tau$ ranked target entities as the candidate entities according to the fused similarity scores. 

\myparagraph{Reward}
%immediate
The reward is a feedback indicating the influence caused by the action. 
Particularly, we want to stimulate the action that maximizes the local similarity and global coherence, and discourage the action that has been taken previously.
As thus, we define the reward at step $i$ as $r_{i+1} = \vec{s^1}(a_i)\cdot\vec{s^2}(a_i) + \vec{s^3}(a_i)$, where $a_i$ represents the chosen action at step $i$.

With these components, a viable solution is to apply the REINFORCE algorithm~\cite{DBLP:journals/ml/Williams92} with policy gradients to maximize the total reward and determine the parameters of the neural network. 
Nevertheless, the drawback of this approach is that the reward can only be calculated after the whole episode is finished, while the situation in each step cannot be well characterized. 
To address this issue, we design a Critic model that approximates the value function of each state and makes an update at each step. 

\myparagraph{Critic Network}
The Critic network comprises two fully connected layers (parameterized by $\eta$). 
At step $i$, the first layer takes as input the state vector, and outputs a hidden vector:
\begin{equation}
\label{eq:critic}
\vec{h_c}({s_i}) = Relu(\vec{W_3}\vec{s_i} + \vec{b_3}).
\end{equation}

Then, the next layer converts the hidden vector into an estimated value:
\begin{equation}
V_\eta(s_{i}) = \vec{W_4}\vec{h_c}({s_i}) + \vec{b_4},
\end{equation}
where $\vec{W_3}, \vec{W_4},\vec{b_3}, \vec{b_4}$ are the learnable parameters.

\myparagraph{Optimization and Learning Procedure}
The training of the Actor and Critic networks are performed separately.
Regarding the Actor, the objective is to obtain the highest total reward, which can be formally defined as:
\begin{equation}
J(\theta) = \mathbb{E} [\sum_{i=0}^{n-1} r_{i+1}|\pi_\theta] = \sum_{i=0}^{n-1} P(s_i,a_i)r_{i+1},
\end{equation}
where $P(s_i,a_i)$ denotes the probability of $(s_i,a_i)$, i.e., choosing action $a_i$ for the state $s_i$.
To optimize the objective, following~\cite{DBLP:conf/icml/MnihBMGLHSK16}, the update of parameter $\theta$ can be written as:
%\begin{equation}
%\nabla_\theta J(\theta) = \sum_{i=0}^{n-1} \nabla_\theta\log\pi_\theta(a_i|s_i)(r_{i+1} + \gamma V_v(s_{i+1}) - V_v(s_{i}))
%\end{equation}
\begin{equation}
\Delta \theta = \alpha\nabla_\theta\log\pi_\theta(a_i|s_i)(r_{i+1} + \gamma V_\eta(s_{i+1}) - V_\eta(s_{i})),
\end{equation}
where $V_\eta(s_{i})$, $V_\eta(s_{i+1})$ correspond to the estimated values generated by the Critic network, $\gamma$ denotes the decay factor, and $r_{i+1} + \gamma V_\eta(s_{i+1}) - V_\eta(s_{i})$ is the Temporal-Difference (TD) error, which can be regarded as a good estimator of 
%the advantage function that plays the role of 
the reward at each step. 
%Gradient ascent is used to update the weights. 
$\alpha$ denotes the learning rate.

As for the Critic, we aim to minimize the mean squared TD error. To achieve the objective, the update of parameter $\eta$ can be written as:
\begin{equation}
\Delta \eta = \beta\nabla_\eta V_\eta(s_i)(r_{i+1} + \gamma V_\eta(s_{i+1}) - V_\eta(s_{i})),
\end{equation}
where $\beta$ denotes the learning rate.

We illustrate the whole learning procedure in Algorithm~\ref{alg:ac} and Figure~\ref{fig:framework}. 
The RL network takes the current state $s_u$ as input, outputs an action $a_u$, which in turn affects the following state $s_{u^\prime}$ and generates a reward $r_u$. Taking the current state $s_u$, the next state $s_{u^\prime}$, and the reward $r_u$ as input, the Critic calculates the TD error and updates its network. The TD error is then fed to the Actor for the network update. This procedure continues until it reaches the final state. 
Notably, the learning sequence is determined according to the highest local similarity score of each source entity. The source entities with higher maximum similarity scores are dealt first.

\begin{algorithm}
	\Input{$E_t^1$:source entities; $E_t^2$: target entities; Adjacency matrices; $\vec {M}$: similarity matrix}
	\Output{A target entity for each source entity}
	\State{Initialize the network parameters}
	\ForEach{epoch}{
		\ForEach{$u \in E_t^1$}{
			\State{Generate the state vector $s_u$, forward it to the Actor}
			\State{Sample an action $a_u$ from $\pi_\theta$}
			\State{Take the action $a_u$ (target entity $v_u$), get the reward $r_u$ and next state $s_{u^\prime}$, forward them to Critic}
			\State{Compute the TD error $\delta = r_{u} + \gamma V_\eta(s_{u^\prime}) - V_\eta(s_{u})$}
			\State{Update the Critic, $\eta \gets \eta +\beta \delta\nabla_\eta V_\eta(s_u)$}
			\State{Update the Actor, $\theta \gets \theta +\alpha \delta\nabla_\theta\log\pi_\theta(a_u|s_u)$}
	}}
	%\Return{$S_a$.}
	\caption{The learning procedure of the A2C network.} \label{alg:ac}
\end{algorithm}

\myparagraph{Preliminary Treatment}
Due to the large number of entities to be aligned, 
%Considering the large scale of entities to be aligned, 
we propose to filter out the source entities that have a high probability of being correctly aligned by merely using the fused similarity matrix $\vec {M}$.
%so as to reduce computational cost of the RL process.
Concretely, for each source entity, if its top-ranked target entity also considers it as the top-ranked source entity, we assume that this entity pair is highly likely to be correct.
Then, we use the rest of the source/target entities as the input to the RL process.

\myparagraph{Remark}
Different from the current collective strategies, which \emph{explicitly} exert the 1-to-1 constraint into the alignment process, the RL framework \emph{implicitly} models the exclusiveness as a component of the state and the reward. 
In this way, although the Actor learns to avoid selecting the already chosen target entities, it still takes into account the other aspects of the current state, i.e., the coherence and local similarity, when making the alignment decision. 
Consequently, the resultant pairing does not strictly follow the 1-to-1 constraint.

\section{Experimental Setting}
\label{exp}
In this section, we describe the experimental settings, including the research questions (Section~\ref{rq}), datasets (Section~\ref{dataset}), parameter settings (Section~\ref{paraset}), evaluation metrics (Section~\ref{metric}), and methods to compare (Section~\ref{compare}). 
The code of \our and the dataset splits can be accessed via \url{https://github.com/DexterZeng/CEAFF}. 

\subsection{Research Questions}
\label{rq}
We attempt to answer the following research questions:
\begin{enumerate}[(RQ1)]
	\item Can \our outperform the state-of-the-art approaches on the EA task? (Section~\ref{rna:rq1})
	
	\item Are the features useful? (Section~\ref{rna:rq5})
	
	\item Is the Bray-Curtis dissimilarity a better distance measure than the commonly used measures, i.e., cosine similarity, Manhattan distance and Euclidean distance? (Section~\ref{rna:rq4})
	
	\item Can the adaptive feature fusion strategy effectively integrate features? (Section~\ref{rna:rq3})

	\item Does the RL-based collective alignment strategy contribute to the performance of \our? (Section~\ref{rna:rq2}) 
	
	\item In which cases does \our fail? and why? (Section~\ref{rna:rq6})
\end{enumerate}

\subsection{Datasets}
\label{dataset}
Three datasets, including eight KG pairs, are used for evaluation:

\dbps. Guo et al.~\cite{DBLP:conf/ijcai/SunHZQ18} established the \dbps dataset. 
They extracted 15 thousand inter-language links (ILLs) in DBpedia~\cite{DBLP:conf/semweb/AuerBKLCI07} with popular entities from English to Chinese, Japanese and French, respectively. These ILLs are also considered as gold standards.

\srp. Guo et al.~\cite{DBLP:conf/icml/GuoSH19} pointed out that KGs in \dbps are too dense and the degree distributions of entities deviate from real-life KGs. 
Therefore, they established a new EA benchmark that follows real-life distribution by using ILLs in DBpedia and reference links among DBpedia, YAGO~\cite{DBLP:conf/www/SuchanekKW07} and Wikidata~\cite{DBLP:journals/cacm/VrandecicK14}. 
They first divided the entities in a KG into several groups by their degrees, and then separately performed random PageRank sampling for each group. 
To guarantee the distributions of the sampled datasets following the original KGs, they used the Kolmogorov-Smirnov (K-S) test to control the difference.
The final evaluation benchmark consists of both cross-lingual and mono-lingual datasets. 
%Each dataset comprises 15,000 aligned entity pairs.

\fb. Zhao et al.~\cite{9174835} noticed that in existing mono-lingual datasets, equivalent entities in different KGs possess identical names from the entity identifiers, which means that a simple comparison of these names can achieve reasonably accurate results. Therefore, they established a new dataset \fb (extracted from DBpedia and Freebase~\cite{DBLP:conf/sigmod/BollackerEPST08}) to mirror the real-life difficulties.

A concise summary of the statistics of the datasets can be found in Table~\ref{tab:data}. 
Note that previous works do not set aside part of the gold standards as the validation set. 
They use 70\% of the gold data as the test set and 30\% as the training set. 
In order to follow a more standard evaluation paradigm, in this work, we use 20\% of the original training set as the validation set for hyper-parameter tuning and used the rest for training. 
The statistics of the new dataset splits are reported in Table~\ref{tab:data}. 
The experiments in this work are all conducted on the new dataset splits.

\begin{table}[htbp]
	\centering
	\caption{The statistics of the evaluation benchmarks. The \#Triples, \#Entities and \#Relations columns indicate the number of triples, entities and relations in each KG, respectively. 
		The \#Align column indicates the total number of gold entity pairs (reference links).
		The \#Test, \#Val and \#Train columns indicate the number of testing, validation, and training entity pairs, respectively.}
	\resizebox{\textwidth}{!}{
		\begin{tabular}{ccccccccc}
			\toprule
			Dataset & KG pairs & \#Triples & \#Entities & \#Relations & \#Align. & \#Test & \#Val & \#Train \\
			\midrule
			\multirow{2}[2]{*}{\dbpsz} & DBpedia(Chinese) & 70,414 & 19,388 & 1,701 & \multirow{2}[2]{*}{15,000} & \multirow{2}[2]{*}{10,500} & \multirow{2}[2]{*}{900} & \multirow{2}[2]{*}{3,600} \\
			& DBpedia(English) & 95,142 & 19,572 & 1,323 &       &       &       &  \\
			\midrule
			\multirow{2}[2]{*}{\dbpsj} & DBpedia(Japanese) & 77,214 & 19,814 & 1,299 & \multirow{2}[2]{*}{15,000} & \multirow{2}[2]{*}{10,500} & \multirow{2}[2]{*}{900} & \multirow{2}[2]{*}{3,600} \\
			& DBpedia(English) & 93,484 & 19,780 & 1,153 &       &       &       &  \\
			\midrule
			\multirow{2}[2]{*}{\dbpsf} & DBpedia(French) & 105,998 & 19,661 & 903   & \multirow{2}[2]{*}{15,000} & \multirow{2}[2]{*}{10,500} & \multirow{2}[2]{*}{900} & \multirow{2}[2]{*}{3,600} \\
			& DBpedia(English) & 115,722 & 19,993 & 1,208 &       &       &       &  \\
			\midrule
			\multirow{2}[2]{*}{\srpf} & DBpedia(English) & 36,508 & 15,000 & 221   & \multirow{2}[2]{*}{15,000} & \multirow{2}[2]{*}{10,500} & \multirow{2}[2]{*}{900} & \multirow{2}[2]{*}{3,600} \\
			& DBpedia(French) & 33,532 & 15,000 & 177   &       &       &       &  \\
			\midrule
			\multirow{2}[2]{*}{\srpd} & DBpedia(English) & 38,281 & 15,000 & 222   & \multirow{2}[2]{*}{15,000} & \multirow{2}[2]{*}{10,500} & \multirow{2}[2]{*}{900} & \multirow{2}[2]{*}{3,600} \\
			& DBpedia(German) & 37,069 & 15,000 & 120   &       &       &       &  \\
			\midrule
			\multirow{2}[2]{*}{\srpw} & DBpedia & 38,421 & 15,000 & 253   & \multirow{2}[2]{*}{15,000} & \multirow{2}[2]{*}{10,500} & \multirow{2}[2]{*}{900} & \multirow{2}[2]{*}{3,600} \\
			& Wikidata & 40,159 & 15,000 & 144   &       &       &       &  \\
			\midrule
			\multirow{2}[2]{*}{\srpy} & DBpedia & 33,571 & 15,000 & 223   & \multirow{2}[2]{*}{15,000} & \multirow{2}[2]{*}{10,500} & \multirow{2}[2]{*}{900} & \multirow{2}[2]{*}{3,600} \\
			& YAGO3 & 34,660 & 15,000 & 30    &       &       &       &  \\
			\midrule
			\multirow{2}[1]{*}{\fb} & DBpedia & 96,414 & 29,861 & 407   & \multirow{2}[1]{*}{25,542} & \multirow{2}[1]{*}{17,880} & \multirow{2}[1]{*}{1,532} & \multirow{2}[1]{*}{6,130} \\
			& Freebase & 111,974 & 25,542 & 882   &       &       &       &  \\
			\bottomrule
		\end{tabular}%
	}
	\label{tab:data}%
\end{table}%

\subsection{Parameter Settings}
\label{paraset}
For learning the {\itshape structural representation}, as it is not the focus of this article, we use the popular parameter settings in existing works~\cite{DBLP:conf/emnlp/WangLLZ18,DBLP:conf/ijcai/WuLF0Y019,cea} and set $d_s$ to 300, $\epsilon$ to 3, the number of training epochs to 300. Five negative examples are generated for each positive pair. 
Regarding the {\itshape entity name representation}, we utilize the pre-trained fastText word embeddings with subword information~\cite{bojanowski2017enriching} as word embeddings and obtain the multilingual word embeddings from MUSE\footnote{\url{https://github.com/facebookresearch/MUSE}}. 
%The fastText embedding models are trained using CBOW with position-weights, in dimension 300, with character n-grams of length 5, a window of size 5 and 10 negatives. 
As for the {\itshape adaptive feature fusion}, we set $\theta_1$ to 0.99, $\theta_2$ to 0.48, which are tuned on the validation set. 
% $\theta_2$ to 0.5
Regarding the {\itshape deep RL model}, we set $\gamma$ to 0.9, $\alpha$ to 0.001, $\beta$ to 0.01, $\tau$ to 10, the dimensionality of the hidden state in Equation~\ref{eq:actor} to 10, the dimensionality of the hidden state in Equation~\ref{eq:critic} to 10. We perform 2 rounds of the preliminary treatment, which will be further discussed in Section~\ref{rna:rq2}. 
These hyper-parameters are tuned on the validation set.

\subsection{Evaluation Metrics}
\label{metric}
We use precision, recall and F1 score as the evaluation metrics.
Precision is computed as the number of correct matches produced by a method divided by the number of matches produced by a method. 
Recall is computed as the number of correct matches produced by a method divided by the total number of correct matches (i.e., gold matches). 
F1 score is computed as the harmonic mean between precision and recall. 
Note that in existing EA datasets, all of the entities in a KG are matchable (i.e, for each entity, there exists an equivalent entity in the other KG), and thus the total number of correct matches equals to the number of entities in a KG.
Besides, \sota EA methods~\cite{DBLP:conf/aaai/SunW0CDZQ20,DBLP:conf/wsdm/MaoWXLW20} generate matches for all the entities in a KG.
Therefore, the number of matches produced by these methods is equal to the number of gold matches, and the values of precision, recall, and F1 score are equal for all current EA methods.  
In the following, unless otherwise specified, we only report the values of precision for these EA methods.	
 
We notice that previous EA methods~\cite{DBLP:conf/emnlp/WangLLZ18,DBLP:conf/ijcai/ChenTYZ17} use Hits@$k$ ($k$=1, 10) and mean reciprocal rank (MRR) as the evaluation metrics. 
For each source entity, they rank the target entities according to the similarity scores in a descending order. 
Hits@$k$ is computed as the number of source entities whose equivalent target entity is ranked in the top-$k$ results divided by the total number of source entities, while MRR characterizes the rank of the ground truth. 
Particularly, Hits@1 is the fraction of correctly aligned source entities among all the aligned source entities, and for \sota EA methods, the Hits@1 value is equal to the precision value, since these methods generate matches for all source entities.

\subsection{Methods to Compare}
\label{compare}
The following state-of-the-art EA methods are used for comparison, which can be divided into two groups---methods merely using structural information, and methods using multi-type information. Specifically, the first group consists of :
\begin{itemize}
	\item \mtranse (2017)~\cite{DBLP:conf/ijcai/ChenTYZ17}: This work uses TransE to learning entity embeddings.
	\item \rsn (2019)~\cite{DBLP:conf/icml/GuoSH19}: This work integrates recurrent neural networks (RNNs) with residual learning to capture the long-term relational dependencies within and between KGs.
	\item \mc (2019)~\cite{DBLP:conf/acl/CaoLLLLC19}: A novel multi-channel graph neural network model is put forward to learn alignment-oriented KG embeddings by robustly encoding two KGs via multiple channels.
	\item \ali (2020)~\cite{DBLP:conf/aaai/SunW0CDZQ20}: This work proposes an EA network which aggregates both direct and distant neighborhood information via attention and gating mechanism.
	\item \kecg (2019)~\cite{DBLP:conf/emnlp/LiCHSLC19}: This paper proposes to jointly learn knowledge embedding model that encodes inner-graph relationships, and cross-graph model that enhances entity embeddings with their neighbors' information.
	\item \itranse (2017)~\cite{DBLP:conf/ijcai/ZhuXLS17}: An iterative training process is used to improve the alignment results.
	\item \bootea (2018)~\cite{DBLP:conf/ijcai/SunHZQ18}: This work devises an alignment-oriented KG embedding framework and a bootstrapping strategy.
	\item \na (2019)~\cite{DBLP:conf/ijcai/ZhuZ0TG19}: This work proposes a neighborhood-aware attentional representation method to learn neighbor-level representation. %by aggregating neighbours' representations with a weighted combination.
	\item \te (2019)~\cite{DBLP:conf/semweb/SunHHCGQ19}: A novel edge-centric embedding model is put forward for EA, which contextualizes relation representations in terms of specific head-tail entity pairs.
	\item \mraea (2020)~\cite{DBLP:conf/wsdm/MaoWXLW20}: This solution directly models cross-lingual entity embeddings by attending over the node's incoming and outgoing neighbors and its connected relations' meta semantics. A bi-directional iterative strategy is used to add newly aligned seeds during training.
\end{itemize}    

The second group includes: 
\begin{itemize}  
	\item \gcn (2018)~\cite{DBLP:conf/emnlp/WangLLZ18}: This work utilizes GCN to generate entity embeddings and combines them with attribute embeddings to align 
	entities in different KGs. 
	%	, which in combination with attribute embeddings, are used to align entities in different KGs.
	\item \jape (2017)~\cite{DBLP:conf/semweb/SunHL17}: In this work, the attributes of entities are harnessed to refine the structural information for alignment. 
	\item \rd (2019)~\cite{DBLP:conf/ijcai/WuLF0Y019}: A relation-aware dual-graph convolutional network is proposed to incorporate relation information via attentive interactions between KG and its dual relation counterpart.
	\item \hgcn (2019)~\cite{DBLP:conf/emnlp/WuLFWZ19}: This work proposes to jointly learn entity and relation representations for EA.
	%	\item \mul~\cite{IJCAI19}: This paper offers a novel framework that unifies the views of entity names, relations and attributes to learn embeddings for mono-lingual EA. It can merely cope with mono-lingual EA.
	\item \gm (2019)~\cite{ACL19}: A local sub-graph of an entity is constructed to represent the entity. Entity name information is harnessed for initializing the overall framework.
	\item \gmehd (2020)~\cite{DBLP:conf/aaai/XuSFSY20}: This work introduces two coordinated reasoning methods to solve the many-to-one problem during the decoding process of EA.
	\item \hman (2019)~\cite{DBLP:conf/emnlp/YangZSLLS19}: This work combines multi-aspect information to learn entity embeddings.
	\item \cea (2020)~\cite{cea}: This work proposes a collective framework that formulates EA as the classic stable matching problem, and solves it with the deferred acceptance algorithm.
	\item \dat (2020)~\cite{dat}: This work conceives a degree-aware co-attention network to effectively fuse available features, so as to improve the performance of entities in tail.  
\end{itemize}

We obtain the results of these methods on the new dataset splits by using the source codes and parameter settings provided by the authors. 
However, the source codes of \gmehd are not available and we adopt its results on \dbps from its original paper, where 30\% of the gold standards are used for training (which equals to the combination of training and validation sets in this work)~\footnote{Theoretically, these results should be higher than its actual results on the new splits (where there are less training data)~\cite{DBLP:conf/wsdm/MaoWXLW20}.}.

We include a string-based heuristic \lev for comparison, which uses Levenshtein distance to measure the distance between entity names and selects for each source entity the closest target entity as the alignment result.

We also report the performance of some variants of \our:
\begin{itemize}    
	\item \oursm: Built on the basic components, this approach replaces the RL-based collective strategy with 1-to-1 constrained stable matching.
	\item \ourc: This approach merely models the \emph{coherence} during the RL process.
	\item \oure: This approach merely models the \emph{exclusiveness} during the RL process.
\end{itemize}

The best results on each dataset are denoted in \textbf{bold}. 
\avg represents the averaged results. 
We use the two-tailed t-test to measure the statistical significance. Significant improvements over \cea are marked with $^\blacktriangle$ ($\alpha=0.05$), significant improvements over \gmehd are marked with $^\vartriangle$ ($\alpha=0.05$), and significant improvements over \oursm are marked with $^\blacklozenge$ ($\alpha=0.05$).

\section{Results and Analyses}
\label{rna}
This section reports the experimental results. 
We first address RQ1 by comparing \our with state-of-the-art EA and ER solutions on the EA datasets (Section~\ref{rna:rq1}). 
Then, by conducting the ablation study and detailed analysis of the components in \our, we answer RQ2, RQ3, RQ4 and RQ5 in Section~\ref{rna:rq5}, Section~\ref{rna:rq4}, Section~\ref{rna:rq3} and Section~\ref{rna:rq2}, respectively. 
Finally, we perform detailed case study and error analysis to answer RQ6 (Section~\ref{rna:rq6}).

\subsection{Performance Comparison}
\label{rna:rq1}
In this subsection, we answer RQ1.  
We first report and discuss the experimental results of \sota EA solutions. 

\begin{table}[htbp]
	\centering
	%	\resizebox{0.9\textwidth}{!}{
	\begin{threeparttable}
		\caption{The precision results on \dbps and \fb.}
		\begin{tabular}{p{2.5cm}p{1.6cm}<{\centering}p{1.6cm}<{\centering}p{1.6cm}<{\centering}p{1.6cm}<{\centering}p{1.6cm}<{\centering}}%{lccccc}
			\toprule
			\multirow{2}[3]{*}{} & \multicolumn{4}{c}{DBP15K}    & \multirow{2}[3]{*}{DBP-FB} \\
			\cmidrule{2-5}          & ZH-EN & JA-EN & FR-EN & AVG   &  \\
			\midrule
			\mtranse & 0.172 & 0.194 & 0.185 & 0.184 & 0.059 \\
			\rsn  &0.519 & 0.520 & 0.565 & 0.535 & 0.241 \\
			\mc & 0.361 & 0.306 & 0.355 & 0.341 & 0.202 \\
			\ali & 0.411 & 0.440 & 0.433 & 0.428 & 0.180 \\
			\kecg  & 0.439 & 0.455 & 0.453 & 0.449 & 0.210 \\
			\itranse & 0.149 & 0.197 & 0.169 & 0.172 & 0.018 \\
			\bootea & 0.574 & 0.555 & 0.592 & 0.574 & 0.223 \\
			\na  & 0.303 & 0.282 & 0.297 & 0.294 & / \\
			\te & 0.580 & 0.542 & 0.573 & 0.565 & 0.253 \\
			\mraea & 0.617 & 0.629 & 0.636 & 0.627 & 0.294 \\
			\midrule
			\gcn & 0.401 & 0.390 & 0.375 & 0.389 & 0.178 \\
			\jape  & 0.368 & 0.327 & 0.273 & 0.323 & 0.065 \\
			\hman  & 0.556 & 0.555 & 0.542 & 0.551 & 0.274 \\
			\rd & 0.692 & 0.751 & 0.876 & 0.773 & 0.660 \\
			\hgcn  & 0.707 & 0.746 & 0.871 & 0.775 & 0.789 \\
			\gm & 0.552 & 0.611 & 0.768 & 0.644 & 0.702 \\
			\gmehd & 0.736 & 0.792 & 0.924 & 0.817 & / \\
			\cea   & 0.776$^{\vartriangle}$ & 0.853$^{\vartriangle}$ & 0.967$^{\vartriangle}$ & 0.865$^{\vartriangle}$ & 0.960 \\
			\midrule
			\lev   & 0.070 & 0.066 & 0.781 & 0.306 & 0.578 \\
			\oursm & 0.806$^{\vartriangle \blacktriangle}$ & 0.866$^{\vartriangle \blacktriangle}$ & 0.971$^{\vartriangle \blacktriangle}$ & 0.881$^{\vartriangle \blacktriangle}$ & 0.962$^{\blacktriangle}$ \\
			\oure & 0.789$^{\vartriangle \blacktriangle}$ & 0.857$^{\vartriangle \blacktriangle}$ & 0.967$^{\vartriangle }$ & 0.871$^{\vartriangle \blacktriangle}$ & 0.957 \\
			\ourc & 0.807$^{\vartriangle \blacktriangle}$ & 0.864$^{\vartriangle \blacktriangle}$ & 0.969$^{\vartriangle \blacktriangle}$ & 0.880$^{\vartriangle \blacktriangle}$ & 0.955 \\
			\our & \textbf{0.811}$^{\vartriangle \blacktriangle \blacklozenge}$ & \textbf{0.868}$^{\vartriangle \blacktriangle \blacklozenge}$ & \textbf{0.972}$^{\vartriangle \blacktriangle \blacklozenge}$ & \textbf{0.884}$^{\vartriangle \blacktriangle \blacklozenge}$ & \textbf{0.963}$^{\blacktriangle \blacklozenge}$ \\
			\bottomrule
		\end{tabular}%
		%	}
		\label{tab:dbp}%
		\begin{tablenotes}
			\footnotesize {
				\item[1] We fail to obtain the result of \na on \fb under our experimental environment, as it requires extremely large amount of memory space. 
				We also fail to obtain the result of \gmehd on \fb since its source code is not available and our implementation cannot reproduce the claimed performance. 
			}
		\end{tablenotes}
	\end{threeparttable}
	%}
\end{table}%

\myparagraph{Results in the First Group}
Solutions in the first group use the structural information for alignment. 
They can be further divided into two categories depending on the use of iterative training strategy. 
The first category includes methods that do not employ the bootstrapping strategy, i.e., \mtranse, \rsn, \mc, \ali and \kecg. 
\mtranse obtains unsatisfactory results as it learns the embeddings in different vector spaces, and suffers from information loss when modeling the transition between different embedding spaces~\cite{DBLP:conf/ijcai/WuLF0Y019}. 
\rsn enhances the performance by taking into account the long-term relational dependencies between entities, which can capture more structural signals for alignment. 
On \dbps, \mc, \ali and \kecg achieve much better results than \mtranse, since they exploit more neighboring information for alignment. 
Concretely, \mc uses a multi-channel graph neural network that captures different levels of structural information. 
\kecg adopts a similar idea by jointly learning entity embeddings that encode both inner-graph relationships and neighboring information.
\ali aggregates both direct and distant neighborhood information via attention and gating mechanism. 
However, they are still outperformed by \rsn. 
Additionally, \mc attains worse performance than \mtranse on \srp, since there are no aligned relations on \srp, where the rule transferring of \mc fails to work. 

Methods in the second category employ bootstrapping strategy to enhance the alignment performance, including \itranse, \bootea, \na, \te and \mraea. 
\bootea achieves better performance than \itranse, since it devises an alignment-oriented KG embedding framework with one-to-one constrained bootstrapping strategy. 
On top of \bootea, \na uses a neighborhood-aware attentional representation model to make better use of the KG structure and learn more comprehensive structural representations.
Nevertheless, using the codes provided by the authors cannot reproduce the results reported in the original paper. 
\te attains promising results, as it employs an edge-centric embedding model to capture structural information, which generates more precise entity embeddings and hence better alignment results. 
%The iterative strategy also helps.
Similarly, \mraea proposes to improve EA performance by modeling relation semantics and employing the bi-directional iterative training strategy, and it attains the best results among the methods that merely use the structural information.

Noteworthily, the overall results on \srp and \fb are worse than \dbps, as the KGs in \dbps are much denser than those in \srp and \fb~\cite{DBLP:conf/icml/GuoSH19,9174835}. 

\begin{table}[htbp]
	\centering
	\begin{threeparttable}
		\caption{The precision results on \srp.}
		\begin{tabular}{p{2cm}p{1.6cm}<{\centering}p{1.6cm}<{\centering}p{1.6cm}<{\centering}p{1.6cm}<{\centering}p{1.6cm}<{\centering}}
			\toprule
			& \srpf & \srpd & \srpw & \srpy & \avg\\
			\midrule
			\mtranse & 0.166 & 0.125 & 0.186 & 0.158 & 0.159 \\
			\rsn  & 0.315 & 0.455 & 0.380 & 0.344 & 0.374 \\
			\mc & 0.114 & 0.225 & 0.130 & 0.159 & 0.157 \\
			\ali & 0.228 & 0.352 & 0.258 & 0.273 & 0.278 \\
			\kecg  & 0.277 & 0.414 & 0.306 & 0.333 & 0.333 \\
			\itranse & 0.084 & 0.088 & 0.069 & 0.064 & 0.076 \\
			\bootea & 0.351 & 0.493 & 0.387 & 0.379 & 0.403 \\
			\na  & 0.178 & 0.306 & 0.187 & 0.200 & 0.218 \\
			\te & 0.355 & 0.501 & 0.388 & 0.379 & 0.406 \\
			\mraea & 0.380  & 0.531 & 0.428 & 0.450  & 0.447 \\
			\midrule
			\gcn & 0.264 & 0.395 & 0.299 & 0.331 & 0.322 \\
			\jape  & 0.236 & 0.263 & 0.217 & 0.189 & 0.226 \\
			\hman  & 0.400 & 0.532 & 0.431 & 0.446 & 0.452 \\
			\rd & 0.580  & 0.676 & 0.989 & 0.993 & 0.810 \\
			\hgcn  & 0.563 & 0.634 & 0.988 & 0.988 & 0.793 \\
			\gm & 0.570 & 0.678 & 0.799 & 0.764 & 0.703 \\
			\dat   & 0.765 & 0.862 & 0.921 & 0.931 & 0.870 \\
			\cea   & 0.966 & 0.977 & 1.000 & 1.000 & 0.986 \\
			\midrule
			\lev   & 0.851 & 0.862 & \textbf{1.000} & \textbf{1.000} & 0.928 \\
			\oursm & 0.966 & 0.979$^{\blacktriangle}$ & \textbf{1.000} & \textbf{1.000} & 0.986 \\
			\oure & 0.964 & 0.979$^{\blacktriangle}$ & \textbf{1.000} & \textbf{1.000} & 0.986 \\
			\ourc & 0.964 & 0.979$^{\blacktriangle}$ & \textbf{1.000} & \textbf{1.000} & 0.986 \\
			\our & \textbf{0.968}$^{\blacktriangle \blacklozenge}$ & \textbf{0.981}$^{\blacktriangle \blacklozenge}$ & \textbf{1.000} & \textbf{1.000} & \textbf{0.987}$^{\blacktriangle \blacklozenge}$ \\
			\bottomrule
		\end{tabular}%
		\label{tab:srp}%
		%	\begin{tablenotes}
		%		\footnotesize {
		%			\item[1] We fail to obtain the results of \gmehd, since there are no available codes and our implementation cannot reproduce the claimed performance. 
		%		}
		%	\end{tablenotes}
	\end{threeparttable}
\end{table}%

\myparagraph{Results in the Second Group}
Taking advantage of the attribute information, \gcn, \jape and \hman outperform \mtranse.
\hman achieves better performance than \jape and \gcn, since (1) it explicitly regards the relation type features as model input; and (2) it employs feedforward neural networks to obtain the embeddings of relations and attributes~\cite{DBLP:conf/emnlp/YangZSLLS19}.

The other methods all exploit the entity name information for alignment, and their results exceed the attribute-enhanced methods.
This verifies the significance of entity names for aligning entities.
Among them, the performance of \rd and \hgcn are close, surpassing \gm.
This is because they employ relations to improve entity embeddings, which has been largely neglected in previous GNN-based EA models.
\dat also attains promising results by improving the alignment performance of long-tail entities.
Both \cea and \gmehd focus on the collective alignment process. While \gmehd views EA as the task assignment problem and employs the Hungarian algorithm, \cea outperforms \gmehd by modeling EA as the stable matching problem and solving it with the deferred acceptance algorithm. 

Notably, it can be observed from Table~\ref{tab:dbp} that the methods using entity names achieve much better results on the datasets with closely-related language pairs (e.g., FR-EN) than those with distantly-related language pairs (e.g., ZH-EN). This shows that the language pairs can influence the use of entity name information and in turn affect the overall alignment performance.

\myparagraph{Results of Our Proposal}
As can be observed from Table~\ref{tab:dbp} and Table~\ref{tab:srp}, our proposal consistently outperforms all other methods on all datasets. 
We attribute the superiority of our model to its four advantages:
\begin{inparaenum} [(1)]
	\item we leverage three representative sources of information, i.e., structural, semantic and string-level features, to offer more comprehensive signals for EA; and 
	\item we adopt the Bray-Curtis dissimilarity to better measure the similarity between entity embeddings; and
	\item we fuse the features with adaptively assigned weights, which can fully take into consideration the strength of each feature; and
	\item we align the source entities collectively via reinforcement learning, which can adequately capture the interdependence between EA decisions.  
\end{inparaenum} 
Particularly, \our achieves better results than the collective alignment methods \cea and \gmehd, and the improvements are statistically significant. 

Moreover, \our achieves superior results than \oursm on most datasets, as it simultaneously models the \emph{exclusiveness} and \emph{coherence} of EA decisions. 
This is also validated by the fact that \our outperforms \oure and \ourc, which merely models the \emph{exclusiveness} and \emph{coherence} during the RL process, respectively.
%, which can mitigate the error propagation caused by the 1-to-1 constraint. A detailed analysis can be found in Section~\ref{error}.
%The resultant paring of \our does not necessarily conform to the 1-to-1 constraint.
Notably, \our advances the precision to 1 on \srpw and \srpy. This is because the entity names in DBpedia, YAGO and Wikidata are nearly identical, where the string-level feature is extremely effective. 
In fact, merely using the Levenshtein distance between entity names (\lev) can already attain ground-truth results on these two datasets. 
%In contrast, although the semantic information is also useful, not all words in entity names can find corresponding entries in the word embeddings (out-of-vocabulary problem), which hence limits its effectiveness.  
In comparison, on \fb, \lev merely attains the precision at 0.578. \our further improves the performance by incorporating the structural and semantic information for alignment.

\begin{table}[htbp]
	\centering
	\begin{threeparttable}
		\caption{Evaluation as the ranking problem on \dbps.}
		\begin{tabular}{lccccccccc}
			\toprule
			& \multicolumn{3}{c}{\dbpsz} & \multicolumn{3}{c}{\dbpsj} & \multicolumn{3}{c}{\dbpsf} \\
			\cmidrule{2-10} & H@1 & H@10 & MRR   & H@1 & H@10 & MRR   & H@1 & H@10 & MRR \\
			\midrule
			\mtranse & 17.2  & 48.4  & 0.274 & 19.4  & 52.8  & 0.304 & 18.5  & 52.7  & 0.297 \\
			\rsn  & 51.9  & 76.4  & 0.606 & 52.0  & 76.1  & 0.605 & 56.5  & 81.8  & 0.653 \\
			\mc & 36.1  & 72.6  & 0.478 & 30.6  & 69.4  & 0.431 & 35.5  & 74.8  & 0.483 \\
			\ali & 41.1  & 67.6  & 0.507 & 44.0  & 69.6  & 0.532 & 43.3  & 72.3  & 0.536 \\
			\kecg  & 43.9  & 80.4  & 0.560 & 45.5  & 82.1  & 0.577 & 45.3  & 82.3  & 0.577 \\
			\itranse & 14.9  & 41.2  & 0.236 & 19.7  & 47.7  & 0.290 & 16.9  & 45.5  & 0.264 \\
			\bootea & 57.4  & 81.7  & 0.657 & 55.5  & 81.4  & 0.641 & 59.2  & 84.6  & 0.678 \\
			\na  & 30.3  & 55.7  & 0.390 & 28.2  & 54.8  & 0.370 & 29.7  & 56.7  & 0.388 \\
			\te & 58.0  & 84.9  & 0.676 & 54.2  & 82.3  & 0.642 & 57.3  & 87.5  & 0.682 \\
			\mraea & 61.7  & \textbf{88.1} & 0.711 & 62.9  & 89.1  & 0.722 & 63.6  & 90.7  & 0.738 \\
			\midrule
			\gcn & 40.1  & 73.7  & 0.516 & 39.0  & 73.3  & 0.507 & 37.5  & 74.7  & 0.500 \\
			\jape  & 36.8  & 70.1  & 0.481 & 32.7  & 65.0  & 0.435 & 27.3  & 62.1  & 0.390 \\
			\hman  & 55.6  & 85.0  & 0.659 & 55.5  & 86.0  & 0.662 & 54.2  & 87.2  & 0.658 \\
			\rd & 69.2  & 83.5  & 0.743 & 75.1  & 87.5  & 0.795 & 87.6  & 95.0  & 0.903 \\
			\hgcn  & 70.7  & 85.3  & 0.759 & 74.6  & 88.0  & 0.794 & 87.1  & 95.3  & 0.901 \\
			\gm & 55.2  & 76.2  & 0.630 & 61.1  & 81.2  & 0.686 & 76.8  & 92.5  & 0.829 \\
			%		\gmehd$^\star$ & 73.6  & /     & /     & 79.2  & /     & /     & 92.4  & /     & / \\
			%		\cea   & 77.6$^{\vartriangle}$  & /     & /     & 85.3$^{\vartriangle}$  & /     & /     & 96.7$^{\vartriangle}$  & /     & / \\
			\midrule
			\our w/o \textsf{C} & \textbf{73.6}  & 86.9  & \textbf{0.785} & \textbf{80.3}  & \textbf{91.1} & \textbf{0.842} & \textbf{93.3}  & \textbf{98.1} & \textbf{0.951} \\
			%		\our & \textbf{81.1}$^{\vartriangle \blacktriangle}$ & /     & /     & \textbf{86.8}$^{\vartriangle \blacktriangle}$ & /     & /     & \textbf{97.2}$^{\vartriangle \blacktriangle}$& /     & / \\
			\bottomrule
		\end{tabular}%
		\label{tab:rank}%
		\begin{tablenotes}
			\footnotesize {
				%			\item[1] $^\star$-marked methods do not release the source codes and we use the results from the original papers (30\% training and 70\% testing). 
				\item[1] H@1 and H@10 represent Hits@1 and Hits@10, respectively. 
				%			Collective alignment methods do not have H@10 and MRR results, which are marked by /.
			}
		\end{tablenotes}
	\end{threeparttable}
\end{table}%

\myparagraph{Evaluation as the Ranking Problem}
For the comprehensiveness of the experiment, following previous works, we consider the EA results in the form of ranked target entity lists, and report the Hits@1, 10 and MRR values in Table~\ref{tab:rank}.
Note that for the collective EA strategies, i.e., \cea, \our and \gmehd, they directly generate the matches rather than the ranked entity lists, and cannot be evaluated by the metrics of the ranking problem, i.e., Hits@1, Hits@10 and MRR. 
Therefore, we do not include them in Table~\ref{tab:rank}. 
Instead, we report the results of \our w/o \textsf{C} where the collective alignment component is removed.
%	, which can be evaluated by the metrics of the ranking problem. 
We leave out the evaluation performance on \srp and \fb in the interest of space. 
It reads from Table~\ref{tab:rank} that, \our w/o \textsf{C} attains the best overall results.

\myparagraph{Comparison with the Collective ER Approach}
As mentioned in Section~\ref{ref}, some collective ER approaches design algorithms to handle graph-structured data. 
Hence, we compare with a representative collective ER approach, \paris~\cite{DBLP:journals/pvldb/SuchanekAS11}. 
% implicitly capture structural information and work well graph-structured data. 
%In order to compare these traditional approaches with \sota EA methods, we run \paris~\cite{DBLP:journals/pvldb/SuchanekAS11}, a collective ER approach, on EA datasets. 
Built on the similarity comparison between literals, \paris devises a probabilistic algorithm to jointly align entities by leveraging the KG structure and attributes~\myfootnote{\paris can also align relations and deal with unmatchable entities, while these are not the focus of EA task.}.

We use precision, recall and F1 score as the evaluation metrics. Different from \sota EA methods, \paris does not necessarily generate a target entity for each source entity, and hence its precision, recall, and F1 score values could be different. 

% Table generated by Excel2LaTeX from sheet 'un'
\begin{table}[htbp]
	\centering
	\caption{Results of \our and \paris on EA datasets. P, R represent precision and recall, respectively.}
	\begin{tabular}{ccccccccccccc}
		\toprule
		\multirow{2}[4]{*}{} & \multicolumn{3}{c}{\dbpsz} & \multicolumn{3}{c}{\dbpsj} & \multicolumn{3}{c}{\dbpsf} & \multicolumn{3}{c}{\srpf} \\
		\cmidrule{2-13}          & P     & R     & F1    & P     & R     & F1    & P     & R     & F1    & P     & R     & F1 \\
		\midrule
		\our & 0.81  & 0.81  & 0.81  & 0.87  & \textbf{0.87} & 0.87  & 0.97  & \textbf{0.97} & \textbf{0.97} & 0.97  & \textbf{0.97} & \textbf{0.97} \\
		\paris & \textbf{0.88} & \textbf{0.87} & \textbf{0.88} & \textbf{0.98} & 0.84  & \textbf{0.90} & \textbf{0.99} & 0.93  & 0.96  & \textbf{0.99} & 0.87  & 0.93 \\
		\midrule
		\multirow{2}[4]{*}{} & \multicolumn{3}{c}{\srpd} & \multicolumn{3}{c}{\srpw} & \multicolumn{3}{c}{\srpy} & \multicolumn{3}{c}{\fb} \\
		\cmidrule{2-13}          & P     & R     & F1    & P     & R     & F1    & P     & R     & F1    & P     & R     & F1 \\
		\midrule
		\our & 0.98  & \textbf{0.98} & \textbf{0.98} & \textbf{1.00} & \textbf{1.00} & \textbf{1.00} & \textbf{1.00} & \textbf{1.00} & \textbf{1.00} & 0.96  & \textbf{0.96} & \textbf{0.96} \\
		\paris & \textbf{0.99} & 0.93  & 0.96  & \textbf{1.00} & \textbf{1.00} & \textbf{1.00} & \textbf{1.00} & \textbf{1.00} & \textbf{1.00} & \textbf{0.98} & 0.56  & 0.71 \\
		\bottomrule
	\end{tabular}%
	\label{tab:conven}%
\end{table}%

As shown in Table~\ref{tab:conven}, overall speaking, \paris achieves high precision while relatively low recall, since it merely generates matches that it believes to be highly confident. 
%In contrast, \our generates for each source entity a target entity, and hence has higher recall and relatively lower precision. 
In terms of F1 score, \our outperforms \paris on most datasets. 
On \dbpsz and \dbpsj, \paris attains better results, as the additional attribute information it considers can provide more useful signals for alignment.

\myparagraph{Running Time Comparison}
We compared the time cost of \our and \cea. 
Since there is randomness in each run of \our, we conducted the experiments on the same dataset split for ten times and reported the averaged time cost, as well as the median and percentile 95 value of the time cost distribution.
It reads from Table~\ref{tab:time} that \cea is more efficient than our proposed RL model on most datasets. This is because: (1) Although \cea has the worst time complexity of $O(n^2)$, empirically, most source entities can find the match in a few rounds given the reasonably accurate similarity matrix; (2) The time cost of \our is influenced by the number of epochs. A larger number of epochs leads to a more stable alignment result, but it also costs more time. 
Notably, \cea is slower than \our on the lager dataset \fb, which suggests that the runtime of \cea grows quickly when the scale of dataset increases. 
Besides, it should be noted that \our is more efficient than most of the other \sota methods (whose runtime costs are shown in Table 6 in~\cite{9174835}). 

\begin{table}[htbp]
	\centering
	\caption{Time costs of \cea and \our (in seconds).}
	\resizebox{\textwidth}{!}{
		\begin{tabular}{cccccccccc}
			\toprule
			\multirow{2}[4]{*}{Method} & \multirow{2}[4]{*}{Metric} & \multicolumn{3}{c}{\dbps} & \multicolumn{4}{c}{\srp}     & \multirow{2}[4]{*}{\fb} \\
			\cmidrule{3-9}          &       & ZH-EN & JA-EN & FR-EN & EN-FR & EN-DE & DBP-WD & DBP-YG &  \\
			\midrule
			\cea   & /     & 285.7 & 187.5 & 109.9 & 102.7 & 110.0   & 112.6 & 94.6  & 615.3 \\
			\midrule
			\multirow{3}[6]{*}{\our} & Mean  & 3757.1 & 3064.1 & 759.0 & 832.8 & 532.6 & 164.9 & 132.9 & 506.6 \\
			\cmidrule{2-10}          & Median & 3828.0 & 3059.2 & 761.0 & 833.6 & 539.6 & 166.3 & 137.0 & 503.9 \\
			\cmidrule{2-10}          & Percentile 95 & 3837.1 & 3079.8 & 762.8 & 847.4 & 540.2 & 172.2 & 138.1 & 512.3 \\
			\bottomrule
		\end{tabular}%
	}
	\label{tab:time}%
\end{table}%

\subsection{Usefulness of Proposed Features}
\label{rna:rq5}
We conduct an ablation study to gain insight into the components of \our, and the results are presented in Table~\ref{tab:ablation}.
%\myfootnote{The ablation results of \oursm are similar, which are omitted in the interest of space.}
\textsf{C} represents the collective alignment strategy, \textsf{AFF} denotes the adaptive feature fusion strategy, $\vec {M^s}, \vec {M^n}, \vec {M^l}$ represent the structural, semantic and string-level features, respectively. 
\our-\textsf{cosine}, \our-\textsf{Manh}, \our-\textsf{Euc} denote replacing the Bray-Curtis dissimilarity with the cosine similarity, Manhattan distance and Euclidean distance, respectively. 
Notably, on \srpw and \srpy, merely using semantic or string-level information can already achieve the precision of 1. Therefore, removing most components does not hurt the performance on these two datasets. 
In the following analysis, unless otherwise specified, we do not discuss the ablation results on \srpw and \srpy. 

\begin{table}[htbp]
	\centering
	\caption{
		The precision results of ablation study and feature analysis.}
	\resizebox{\textwidth}{!}{
		\begin{tabular}{cccccccccc}
			\toprule
			& \multicolumn{3}{c}{\dbps} & \multicolumn{4}{c}{\srp}     & \multirow{2}[4]{*}{\fb} & \multirow{2}[4]{*}{\avg} \\
			\cmidrule{2-8}          & ZH-EN & JA-EN & FR-EN & EN-FR & EN-DE & DBP-WD & DBP-YG &       &  \\
			\midrule
			\our & \textbf{0.811} & \textbf{0.868} & \textbf{0.972} & \textbf{0.968} & \textbf{0.981} & \textbf{1.000} & \textbf{1.000} & \textbf{0.963} & \textbf{0.945} \\
			w/o \textsf{C} & 0.736 & 0.803 & 0.933 & 0.931 & 0.946 & \textbf{1.000} & \textbf{1.000} & 0.814 & 0.895 \\
			w/o \textsf{AFF} & 0.800 & \textbf{0.868} & 0.971 & 0.966 & 0.980 & \textbf{1.000} & \textbf{1.000} & 0.957 & 0.943 \\
			w/o $\vec {M^s}$ & 0.657 & 0.748 & 0.938 & 0.935 & 0.955 & \textbf{1.000} & \textbf{1.000} & 0.751 & 0.873 \\
			w/o $\vec {M^n}$ & 0.471 & 0.507 & 0.950 & 0.957 & 0.972 & \textbf{1.000} & \textbf{1.000} & 0.913 & 0.846 \\
			w/o $\vec {M^l}$ & 0.801 & 0.860 & 0.939 & 0.742 & 0.839 & \textbf{1.000} & \textbf{1.000} & 0.901 & 0.885 \\
			w/o $\theta_1, \theta_2$ & 0.799 & 0.863 & \textbf{0.972} & 0.967 & \textbf{0.981} & \textbf{1.000} & \textbf{1.000} & 0.958 & 0.943 \\
			\midrule
			\our-\textsf{cosine} & 0.784 & 0.854 & 0.966 & 0.967 & 0.978 & \textbf{1.000} & \textbf{1.000} & 0.956 & 0.938 \\
			\our-\textsf{Manh} & 0.733 & 0.794 & 0.953 & 0.941 & 0.962 & \textbf{1.000} & \textbf{1.000} & 0.919 & 0.907 \\
			\our-\textsf{Euc} & 0.787 & 0.850 & 0.965 & 0.963 & 0.972 & \textbf{1.000} & \textbf{1.000} & 0.939 & 0.934 \\
			\midrule
			\textsf{AFF}   & 0.736 & 0.803 & 0.933 & 0.931 & 0.946 & \textbf{1.000} & \textbf{1.000} & 0.814 & 0.895 \\
			\textsf{LR}    & 0.738 & 0.801 & 0.937 & 0.919 & 0.934 & \textbf{1.000} & \textbf{1.000} & 0.812 & 0.893 \\
			\textsf{LM}    & 0.665 & 0.727 & 0.915 & 0.909 & 0.932 & \textbf{1.000} & \textbf{1.000} & 0.798 & 0.868 \\
			\bottomrule
		\end{tabular}%
	}
	\label{tab:ablation}%
\end{table}%

To address RQ2, in this subsection, we examine the usefulness of our proposed features (\our vs. \our\textsf{-$\vec {M^s}$, $\vec {M^n}$, $\vec {M^l}$}). 
%Removing structural or semantic features brings performance drop on cross-lingual datasets, whereas it does not hurt the results on mono-lingual datasets. 
% In comparison, eliminating string-level feature will bring consistent 
As shown in Table~\ref{tab:ablation}, removing the structural information leads to lower alignment results. %, showcasing its stable effectiveness across all language pairs. 
Besides, the semantic information plays a more important role on KGs with distantly-related language pairs, e.g., \dbpsz, while the string-level feature is significant for aligning KGs with closely-related language pairs, e.g., \srpf. 

\subsection{Influence of Distance Measures}
\label{rna:rq4}
In this subsection, we address RQ3. 
As can be observed from Table~\ref{tab:ablation}, replacing the Bray-Curtis dissimilarity with other distance measures brings down the performance (\our vs. \our-\textsf{cosine}, \our-\textsf{Manh}, \our-\textsf{Euc}), which validates that an appropriate distance measure can better capture the similarity between entities. 

Then, we remove the influence of the collective alignment and adaptive feature fusion strategies, and directly compare the performance of these four distance measures. 
As depicted in Table~\ref{tab:dm}, \textbf{Structure} denotes merely using structural embeddings for alignment, \textbf{Name} denotes merely using entity name embeddings for alignment, while \textbf{Comb.} refers to fusing structural, semantic and string information with equal weights. 
The performance of solely using string similarity is omitted, since it is not influenced by distance measures. 

% Table generated by Excel2LaTeX from sheet 'feature-comb'
\begin{table}[htbp]
	\centering
	\caption{The precision results of alignment by using different distance measures. The results on \srpy and \srpd are omitted in the interest of space, which are similar to \srpw and \srpf, respectively.}
	\resizebox{\textwidth}{!}{
		\begin{tabular}{cccccccccc}
			\toprule
			& Structure & Name  & Comb. & Structure & Name  & Comb. & Structure & Name  & Comb. \\
			\midrule
			& \multicolumn{3}{c}{ZH-EN} & \multicolumn{3}{c}{JA-EN} & \multicolumn{3}{c}{FR-EN} \\
			\midrule
			BC    & \textbf{0.369} & \textbf{0.597} & \textbf{0.730} & \textbf{0.381} & \textbf{0.669} & \textbf{0.801} & \textbf{0.369} & \textbf{0.822} & \textbf{0.930} \\
			Cosine & 0.333 & 0.584 & 0.707 & 0.343 & 0.656 & 0.778 & 0.325 & 0.811 & 0.927 \\
			Manhattan & 0.355 & 0.596 & 0.680 & 0.356 & 0.664 & 0.740 & 0.355 & 0.821 & 0.911 \\
			Euclidean & 0.354 & 0.584 & 0.719 & 0.353 & 0.656 & 0.781 & 0.346 & 0.811 & 0.928 \\
			\midrule
			& \multicolumn{3}{c}{EN-FR} & \multicolumn{3}{c}{DBP-WD} & \multicolumn{3}{c}{DBP-FB} \\
			\midrule
			BC    & \textbf{0.242} & 0.524 & \textbf{0.930} & \textbf{0.270} & \textbf{1.000} & \textbf{1.000} & \textbf{0.171} & 0.568 & \textbf{0.810} \\
			Cosine & 0.196 & \textbf{0.530} & \textbf{0.930} & 0.264 & \textbf{1.000} & \textbf{1.000} & 0.149 & \textbf{0.583} & 0.775 \\
			Manhattan & 0.224 & 0.515 & 0.904 & 0.253 & \textbf{1.000} & \textbf{1.000} & 0.158 & \textbf{0.583} & 0.752 \\
			Euclidean & 0.222 & 0.426 & 0.920 & 0.252 & \textbf{1.000} & \textbf{1.000} & 0.156 & \textbf{0.583} & 0.807 \\
			\bottomrule
		\end{tabular}%
	}
	\label{tab:dm}%
\end{table}%

It shows in Table~\ref{tab:dm} that, using the Bray-Curtis dissimilarity (BC) brings the best performance on all datasets in terms of \textbf{Structure} and \textbf{Comb.}. 
Regarding the \textbf{Name} category, the Bray-Curtis dissimilarity also attains the best results on most datasets. 
This further verifies the effectiveness of this distance measure.
% in terms of solely using structure information and combining different features for alignment
%statistically significant improvements (P < 0.05) were marked by

\subsection{Effectiveness of the Adaptive Feature Fusion Strategy}
\label{rna:rq3}
In this subsection, we proceed to answer RQ4 and
%Specifically, we then 
examine the contribution of the adaptive feature fusion strategy (\our vs. \our\textsf{w/o AFF} in Table~\ref{tab:ablation}). 
Specifically, we replace the dynamic weight assignment with fixed weights, i.e., the same weight for each feature. 
As can be observed from Table~\ref{tab:ablation}, overall speaking, using the adaptive feature fusion strategy improves the results of using the fixed weighting, validating its usefulness. 
It is also noted that the increment is not very significant. 
This is because, by treating different features equally, the averaged weighting is the safest and the most common approach for feature fusion. 
Although our proposed adaptive feature fusion strategy can optimize the assignment of weights adaptively, it cannot change the input features, and hence cannot bring substantial improvement. 

\myparagraph{Thresholds $\theta_1, \theta_2$ in Adaptive Feature Fusion}
As mentioned in Section~\ref{ff}, for a correspondence with very large similarity score, i.e., exceeding $\theta_1$, we set its weight to a small value $\theta_2$. 
To examine the usefulness of this strategy, we report the results after removing this setting in Table~\ref{tab:ablation}. Without this setting, the performance drops on most datasets, verifying its effectiveness. 
%Theoretically speaking, after applying this setting, the features that are very effective would not be assigned with extremely large values, and less effective features can always contribute to the final EA decisions. 

\myparagraph{The Learning-based Fusion Strategy}
Our adaptive feature fusion strategy (\textsf{AFF}) can dynamically determine the weights of features without training data. 
For the comprehensiveness of the experiment, we compare with stronger baselines with learnable parameters in terms of aggregating features for alignment. 
We first adopt the Logistic Regression algorithm (\textsf{LR}) to determine the weights of features by casting the alignment to the classification problem, i.e., labeling correct EA pairs with ``1''s and false pairs with ``0''s. 
We use the learned weights to combine features and rank the target entities according to the fused similarity matrix. 
Note that we do not employ the collective alignment strategy so as to exclude its influence on the results. 
Besides, we also adopt a learning-to-rank algorithm, \textsf{LambdaMART} (\textsf{LM})~\cite{burges2010ranknet}, to learn how to directly aggregate features and rank the target entities without generating the weights of features. 

To construct a more useful training set, for each positive pair, we generate the negative samples by replacing the gold target entity with an incorrect target entity from the set of top-10 ranked target entities produced by solely using the structural, semantic or string-level feature, respectively. 
In this way, the learned models can be more discriminative compared with using randomly selected negative samples. 
Notably, since the original training set is used as the seed to unify the individual structural embedding spaces, the entities in the entity pairs of the training set have very high structural similarity. 
Therefore, it is inappropriate to still use the original training set to learn how to combine the features and rank the target entities (in which case the structural feature would be ``favored'' and considered as extremely useful). 
Instead, we use the validation set to train the models. 
We use the default hyper-parameters of these two models since there is no extra data for parameter optimization. 
For \textsf{LM}, we adopt Precision@1 as the metric to optimize on the training data. 	

The results are reported in Table~\ref{tab:ablation}. It reads that, the performance of \textsf{LM} is not promising, which, to a large extent, can be attributed to the lack of training data (i.e., only 900 pairs are used for training, while 10,500 for testing). 
The effectiveness of \textsf{LR} is also constrained by the lack of training data. 
Nevertheless, it achieves better results than \textsf{LM} and even outperforms \textsf{AFF} on \dbpsz and \dbpsf.  
Overall speaking, the performance of \textsf{AFF} is better than \textsf{LR} and \textsf{LM}, and it does not need training data.

\subsection{Effectiveness of the RL-based Collective Strategy}
\label{rna:rq2}
In this subsection, we seek to answer RQ5.
%	, i.e., the RL-based collective alignment strategy.} 
%In this subsection, we examine the effectiveness of the RL-based collective alignment strategy. 
After removing the collective strategy (\our vs. \our\textsf{w/o C}), the performance drops on all cross-lingual datasets and \fb, revealing the significance of considering the interdependence between EA decisions. 
Moreover, it observes from Table~\ref{tab:dbp} and Table~\ref{tab:srp} that, simultaneously modeling the \emph{exclusiveness} and \emph{coherence} under the RL framework consistently outperforms the alternative strategies (\our vs. \oursm, \oure, \ourc).

\begin{table}[htbp]
	\centering
	\caption{Analysis of the collective alignment constraints. \# \emph{MulSE} denotes the number of source entities that are matched to the same target entities. \# \emph{MulTE} denotes the number of target entities that are assigned multiple times. The analysis on \srpw and \srpy are omitted as the precision is 1 for all methods.}
	\resizebox{0.8\textwidth}{!}{
		\begin{tabular}{clcccccc}
			\toprule
			\multirow{2}[4]{*}{Metric} & \multicolumn{1}{c}{\multirow{2}[4]{*}{Method}} & \multicolumn{3}{c}{\dbps} & \multicolumn{2}{c}{\srp} & \multirow{2}[4]{*}{\fb} \\
			\cmidrule{3-7}          &       & ZH-EN & JA-EN & FR-EN & EN-FR & EN-DE &  \\
			\midrule
			\multirow{3}[2]{*}{Precision} & \multicolumn{1}{c}{\our} & 0.811 & 0.868 & 0.972 & 0.968 & 0.981 & 0.963 \\
			& \multicolumn{1}{c}{\oursm} & 0.806 & 0.866 & 0.971 & 0.966 & 0.979 & 0.962 \\
			& \our w/o \textsf{C} & 0.736 & 0.803 & 0.933 & 0.931 & 0.946 & 0.814 \\
			\midrule
			\multirow{3}[2]{*}{\# \emph{MulSE}} & \multicolumn{1}{c}{\our} & 1116  & 768   & 176   & 173   & 97    & 307 \\
			& \multicolumn{1}{c}{\oursm} & 0     & 0     & 0     & 0     & 0     & 0 \\
			& \our w/o \textsf{C} & 3778  & 3061  & 1142  & 1157  & 983   & 4308 \\
			\midrule
			\multirow{3}[2]{*}{\# \emph{MulTE}} & \multicolumn{1}{c}{\our} & 423   & 307   & 87    & 77    & 43    & 134 \\
			& \multicolumn{1}{c}{\oursm} & 0     & 0     & 0     & 0     & 0     & 0 \\
			& \our w/o \textsf{C} & 1441  & 1272  & 483   & 490   & 437   & 1130 \\
			\bottomrule
		\end{tabular}%
	}
	\label{tab:mul}%
\end{table}%

\myparagraph{Analysis of Collective Alignment Constraints}
We make a detailed analysis of the collective alignment constraints. 
It can be seen from Table~\ref{tab:mul} that, without exerting any constraint on the alignment results, the performance of \our w/o \textsf{C} is much worse than that of the collective alignment methods \our and \oursm. 
Besides, in its matching results, there are many source entities that are aligned to the same target entities, e.g., around 30\% of the source entities on \dbpsz and \dbpsj. 
This could restrain its performance since the ground-truth results in current EA benchmarks are 1-to-1 mappings.   

By exerting the 1-to-1 constraint, \oursm improves the performance, and produces a 1-to-1 mapping result (there are no \emph{MulSE} and \emph{MulTE}). Nevertheless, as illustrated in Example~\ref{exa:121}, the 1-to-1 constraint could cause the error propagation issue. 
In this work, we use the exclusiveness constraint to relax the 1-to-1 constraint and use the coherence constraint to keep the alignment decisions coherent. The resulting model \our attains better performance than \oursm and \our w/o \textsf{C}. 
Additionally, as shown in Table~\ref{tab:mul}, employing the exclusiveness constraint allows a small number of source entities to match with the same target entities, which can partially mitigate the error propagation issue.

\myparagraph{Influence of Preliminary Treatment}
We further analyze the effectiveness of the preliminary treatment of the RL model, i.e., filtering out the source entities that can be correctly aligned by merely using the fused similarity scores. 
We first report the number of source entities that are matched by the preliminary treatment, as well as the percentage of the correctly aligned ones, in Table~\ref{tab:pt}. 
The results prove that our preliminary treatment strategy can filter out many source entities, among which most are correctly aligned. 

\begin{table}[htbp]
	\centering
	\caption{The number of source entities that are matched by the preliminary treatment (two rounds) and the percentage of correctly aligned entities (PoC).}
	\resizebox{0.8\textwidth}{!}{
		\begin{tabular}{ccccccccc}
			\toprule
			\multirow{2}[2]{*}{} & \multicolumn{3}{c}{\dbps} & \multicolumn{4}{c}{\srp}     & \multirow{2}[2]{*}{\fb} \\
			\cmidrule{2-8}          & ZH-EN & JA-EN & FR-EN & EN-FR & EN-DE & DBP-WD & DBP-YG &  \\
			\midrule
			Number & 8013  & 8783  & 10076 & 10017 & 10198 & 10500 & 10500 & 16781 \\
			PoC & 94.4\%  & 96.0\%  & 99.1\%  & 99.2\%  & 99.4\%  & 100.0\% & 100.0\% & 98.3\% \\
			\bottomrule
		\end{tabular}%
	}
	\label{tab:pt}
\end{table}%

Then, we examine: (1) whether the preliminary treatment contributes to the performance of the RL-based alignment model; and (2) the appropriate rounds for conducting the filtering process. 

\begin{figure}[htbp]
	\centering
	\includegraphics[width=0.5\linewidth]{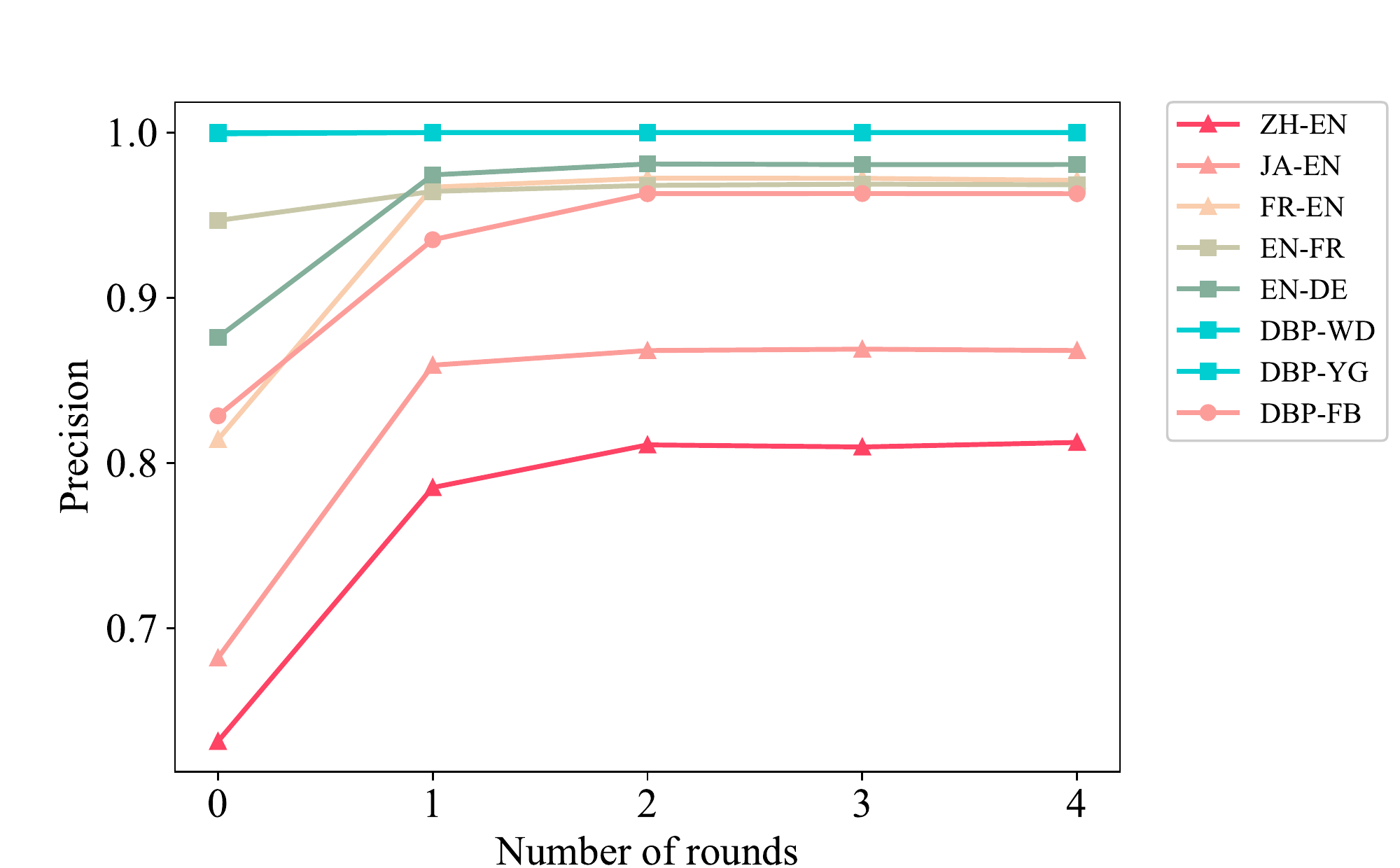}
	\caption{Precision of \our with different rounds of the preliminary treatment.}
	\label{fig:rl}
\end{figure}

Figure~\ref{fig:rl} indicates that, this filtering process does improve the alignment performance, because: (1) it prevents the source entities that can already be confidently aligned using similarity scores from being misled by the inaccurate signals during the RL process; and (2) it reduces the learning steps in each episode and leads to faster convergence; and (3) the confident matches detected by the preliminary treatment can provide more accurate coherence information for aligning the rest of the source entities. 
Additionally, applying this treatment for 2 rounds is enough, since the filtering process cannot generate 100\% correct confident matches on most datasets. 
More rounds of the preliminary treatment increase the errors, which in turn could hurt the overall performance.

\subsection{Error Analysis and Case Study}
\label{rna:rq6}
Finally, we turn to RQ6. We perform error analysis and case study 
%of \our to demonstrate the contribution of each module, and 
to examine the cases where \our falls short.

Firstly, we analyze the change of the error rate after adding our proposed components. 
Take \dbpsz for example. Solely using structural information leads to 63.1\% error rate of the precision.
Incorporating the entity name to complement the structural information reduces the overall error rate to 27\%.
After applying the adaptive feature fusion strategy, the error rate drops to 26.4\%.
On top of it, using the RL framework to collectively align entities leads to an error rate of 18.9\%. 
This implies that, all the components in our model contribute to the alignment performance.
% and lower the percentage of incorrect matches.

Nevertheless, we note that \our still fails to generate correct target entities for some source entities, especially on \dbpsz and \dbpsj. 
Therefore, we carefully analyze the erroneous matches and broadly divide them into five categories:

\myparagraph{Category 1} This group of cases are misled by the structural information, e.g., the match (\textit{Saison 26 des Simpson}, \textit{The Simpsons (season 23)}). 
Using the entity name information, it is easy to align the correct target entity \textit{The Simpsons (season 26)} to the source entity. 
Nevertheless, the target entity \textit{The Simpsons (season 23)} shares more common neighboring entities with the source entity \textit{Saison 26 des Simpson}, and hence is falsely considered as the corresponding target entity.

\myparagraph{Category 2} This group of cases are misled by the textual information, e.g., the match (\textit{Sion (Valais)}, \textit{Valais}). By merely using the structural information, the correct target entity \textit{Sion,\_Switzerland} can be aligned to \textit{Sion (Valais)}. However, the high textual similarity between (\textit{Sion (Valais)}, \textit{Valais}) leads to the wrong match, despite of their structural disparity.

\myparagraph{Category 3} Cases in this group are misled by the feature fusion strategy, e.g., the match (\textit{H\'{e}rouxville}, \textit{H\'{e}rault}). It is noted that using structural information or entity name information can find the correct target entity \textit{H\'{e}rouxville,\_Quebec}, while combining these features leads to the wrong result. 

\myparagraph{Category 4} The collective alignment strategy is responsible for this category of errors, e.g., the match (\textit{Grande Ourse}, \textit{Landrienne,\_Quebec}). 
Based on the fused similarity matrix, generating the results independently can find the correct target entity \textit{Ursa Major} for the source entity. 
Nevertheless, the target entity \textit{Ursa Major} has a higher similarity score with another source entity \textit{Bouvier (constellation)}. As thus, when aligning entities collectively, the \emph{exclusiveness} constraint prevents \textit{Grande Ourse} from aligning \textit{Ursa Major}. 

\myparagraph{Category 5} Cases in this group cannot be tackled by any module in our model, e.g., the match (\textit{Yolande de Hongrie (reine d'Aragon)}, \textit{Constance of Sicily,\_Queen of Aragon}). 
This might be ascribed to the fact that the correct target entity \textit{Violant of Hungary} neither has a similar name, nor shares similar structural information, with the source entity.

% Table generated by Excel2LaTeX from sheet 'error'
\begin{table}[htbp]
	\centering
	\caption{The percentages of different types of errors made by \our.}
	\begin{tabular}{cccccccc}
		\toprule
		\multirow{2}[4]{*}{} & \multicolumn{3}{c}{\dbps} & \multicolumn{2}{c}{\srp} & \multirow{2}[4]{*}{\fb} & \multirow{2}[4]{*}{\avg} \\
		\cmidrule{2-6}          & ZH-EN & JA-EN & FR-EN & EN-FR & EN-DE &       &  \\
		\midrule
		Error Rate & 18.9\% & 13.2\% & 2.8\% & 3.2\% & 1.9\% & 3.7\% & 7.3\% \\
		\midrule
		Category 1 & 2.7\% & 4.7\% & 15.6\% & 14.8\% & 12.8\% & 11.4\% & 10.3\% \\
		Category 2 & 17.8\% & 16.4\% & 15.6\% & 8.3\% & 10.7\% & 3.2\% & 12.0\% \\
		Category 3 & 0.1\% & 0.1\% & 6.3\% & 0.6\% & 0.0\% & 0.0\% & 1.2\% \\
		Category 4 & 11.1\% & 14.1\% & 18.9\% & 15.1\% & 17.3\% & 15.6\% & 15.4\% \\
		Category 5 & 68.3\% & 64.7\% & 43.6\% & 61.2\% & 59.2\% & 69.8\% & 61.1\% \\
		\bottomrule
	\end{tabular}%
	\label{tab:errorana}%
\end{table}%

We report the percentages of these categories of errors in Table~\ref{tab:errorana}~\myfootnote{The results on \srpw and \srpy are omitted since they do not generate erroneous matches.}. 
Next, we analyze the errors and suggest some possible research directions in the future. 

It can be observed from Table~\ref{tab:errorana} that, Category 1 and Category 2 account for 10.3\% and 12\% on average, respectively. This shows that there is still room for improving the feature encoders to learn better representations. 
The feature fusion strategy (Category 3), however, brings few errors. 
15.4\% of the incorrect matches are caused by the collective alignment strategy (Category 4). Admittedly, aligning entities jointly can better model the correlations between EA decisions, and hence significantly improves the alignment results. 
Nevertheless, an erroneous match can trigger error propagation and leads to more erroneous matches, as shown in the example of Category 4. 
Therefore, more advanced collective alignment strategies could be devised to further mitigate this issue.  
Notably, Category 5 takes the largest share, revealing that the majority of errors are not generated by the components of our model. 
However, these errors might be avoided by mining more useful features or designing more advanced approaches to exploit available features. 
%, which will be left as future work. 
%The error analysis and case study also inspire us to investigate some potential improvements in the future, which will be described in Section~\ref{conclude}.

\section{Conclusion}
\label{conclude}
When making EA decisions, current EA solutions treat entities separately, or fail to adequately model the interdependence among entities. 
To fill in this gap, we frame EA as the sequence decision process and devise a deep RL model to capture both the exclusiveness and coherence of EA decisions.
Besides, we put forward an adaptive feature fusion strategy to aggregate multiple features and provide more accurate inputs to the RL framework. 
Compared with state-of-the art approaches, our proposal achieves consistently better results, and the ablation study also verifies the usefulness of each component. 
More concretely, the results show that: (1) aligning entities collectively using reinforcement learning can sufficiently capture the interdependence between EA decisions; and (2) dynamically fusing the features with adaptively assigned weights can better take into consideration the strength of each feature compared with equal weights; and (3) compared with existing distance measures, the Bray-Curtis dissimilarity is a better measure to characterize the distance between entity embeddings.

%We conduct error analysis and further experiments to gain insights into our proposal.

We will explore the following directions in the future: (1) devising more advanced feature encoders that can better exploit available features for alignment; and (2) designing collective alignment algorithms that can further mitigate the error propagation caused by the erroneous matches; and (3) establishing a more challenging (mono-lingual) EA benchmark.

\bibliographystyle{ACM-Reference-Format}
\bibliography{sample-base}

\clearpage
\appendix
\section*{APPENDIX}

\section{Further Analysis of the String Feature} % This is Appendix A
This appendix provides more detailed analysis of the string feature. 
First, we report the average, median, percentile-10, and percentile-90 Levenshtein distance between the names of the gold entity pairs in each dataset in Table~\ref{tab:dis}. 
It can be observed that the string feature is extremely useful on \srpw and \srpy. This is because these two datasets are extracted from DBpedia, Wikidata and YAGO, where equivalent entities in different KGs possess identical labels, and a simple comparison of these labels can achieve ground-truth results~\cite{9174835}. 
%In comparison, the distribution of the Levenshtein distance between the gold entity pairs in another mono-lingual dataset \fb is relatively more normal. 

%Besides, the string feature is also effective on mono-lingual datasets (\fb) and cross-lingual datasets 

\begin{table}[htbp]
	\centering
	\caption{Statistics of the Levenshtein distance between the names of the gold entity pairs in each dataset.}
	\resizebox{0.8\textwidth}{!}{
		\begin{tabular}{ccccccccc}
			\toprule
			\multirow{2}[2]{*}{} & \multicolumn{3}{c}{\dbps} & \multicolumn{4}{c}{\srp}     & \multirow{2}[2]{*}{\fb} \\
			\cmidrule{2-8}          & ZH-EN & JA-EN & FR-EN & EN-FR & EN-DE & DBP-WD & DBP-YG &  \\
			\midrule
			Average   & 13.8  & 14.5  & 5.2   & 3.3   & 3.1   & 0     & 0     & 4.8 \\
			Median & 13    & 14    & 0     & 0     & 0     & 0     & 0     & 4 \\
			Percentile-90 & 23    & 24    & 16    & 13    & 12    & 0     & 0     & 12 \\
			Percentile-10 & 6     & 7     & 0     & 0     & 0     & 0     & 0     & 0 \\
			\bottomrule
		\end{tabular}%
	}
	\label{tab:dis}%
\end{table}%

We further investigate into the string feature by creating a new dataset (\texttt{WD\_IMDB}) with KGs from independent sources (Wikidata and IMDB) and comparing the distribution of the Levenshtein distance between the entity names of gold entity pairs on \texttt{DBP\_WD} and \texttt{WD\_IMDB}. 

\begin{figure*}[htbp]
	\centering
	\begin{minipage}{.4\textwidth}
		\includegraphics[width=6cm]{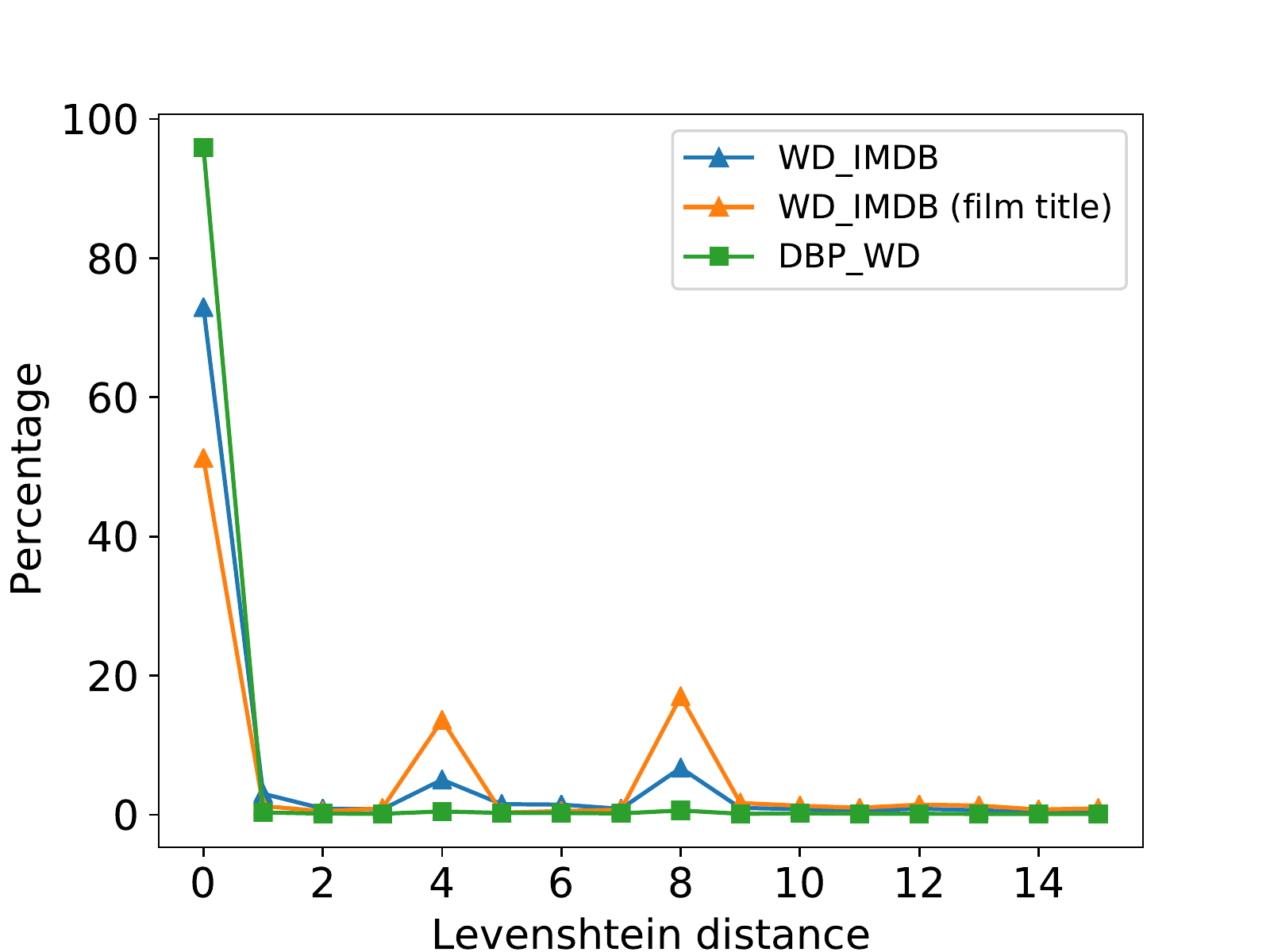}
		\captionof{subfigure}{On \textsf{DBP\_WD} and \textsf{WD\_IMDB}}
		%		\label{subfig:rating}
	\end{minipage}
	\goodgap \goodgap
	\begin{minipage}{.4\textwidth}
		\includegraphics[width=6cm]{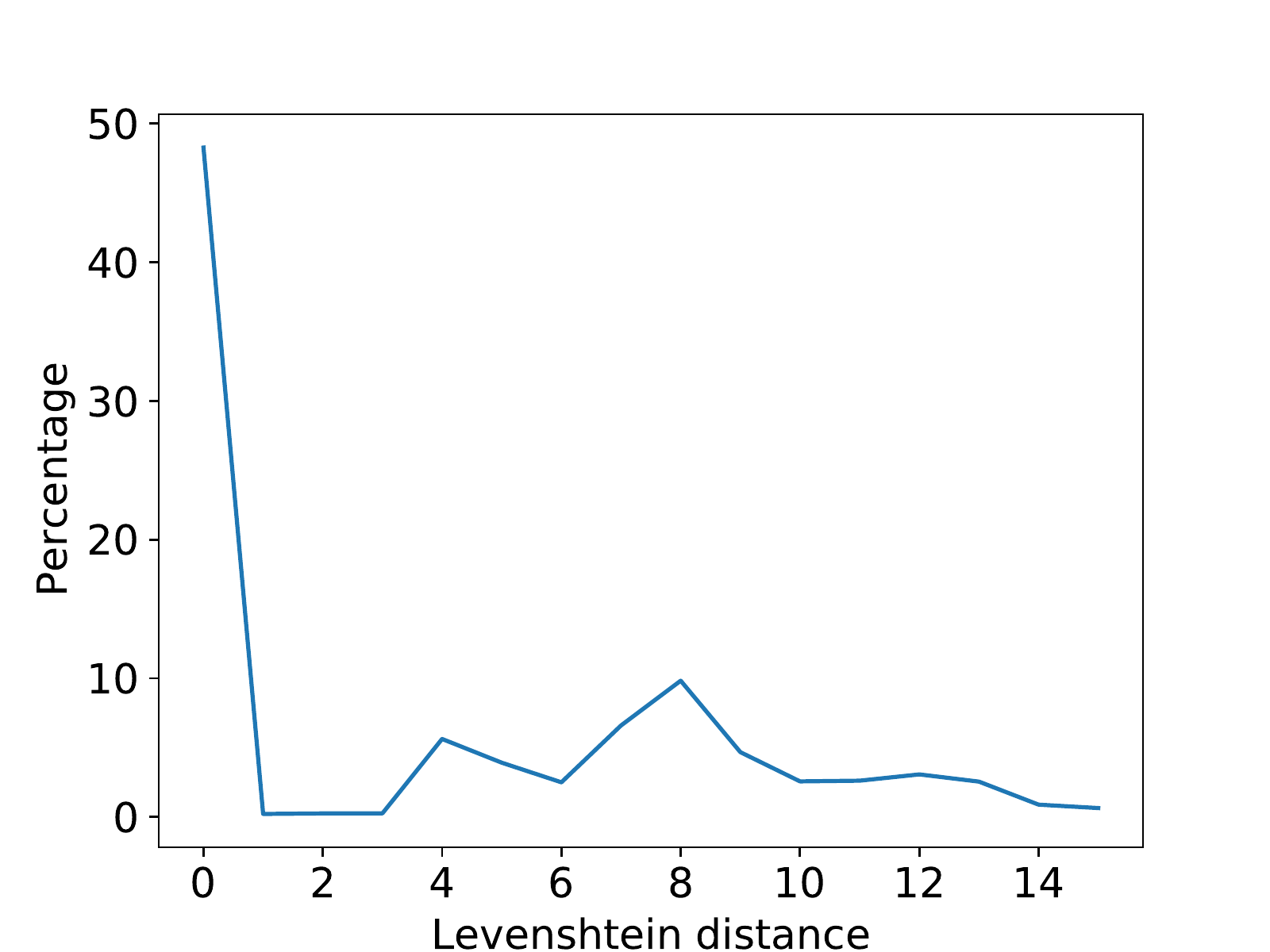}
		\captionof{subfigure}{On \textsf{DBP\_FB}}
		%		\label{subfig:lvsm}	
	\end{minipage}
	\caption{Distribution of the Levenshtein distance between the names of gold entity pairs.}\label{fig:f}
\end{figure*}

More specifically, we created the Wikidata-IMDB KG pair, \texttt{WD\_IMDB}, as well as the \texttt{WD\_IMDB (film title)} dataset that only comprises film entities. 
Figure~\ref{fig:f}-(a) shows the the distribution of the Levenshtein distance between the names of the gold entity pairs in \texttt{WD\_IMDB}, \texttt{WD\_IMDB (film title)} and \texttt{DBP\_WD}. 
It reads that on \texttt{WD\_IMDB}, the distribution of Levenshtein distance is relatively more ``normal'' than that on \texttt{DBP\_WD}. 
Besides, there is a less portion of entities that have exactly the same labels on \texttt{WD\_IMDB (film title)} compared with \texttt{WD\_IMDB}.
This is because many entities in \texttt{WD\_IMDB} are film-related persons, and the names of persons tend to be the same in different KGs, while the titles of film entities are more likely to be described differently. 

In summary, it can be concluded that the name information is indeed causing overfitting issue on DBpedia, Wikidata and YAGO. 
As a consequence, we adopt a recently constructed dataset \fb, which also mitigates the overfitting issue of the string feature, to evaluate \sota methods.  
The distribution of the Levenshtein distance over \fb is shown in Figure~\ref{fig:f}-(b), which is similar to \texttt{WD\_IMDB (film title)}.
We reckon \fb is a more appropriate dataset for EA compared with \texttt{WD\_IMDB}, since IMDB merely contains entities in a few types (mostly films and persons) and is domain-specific, while Freebase is a more general KG. 
In consequence, we include the \fb dataset in the experiment and evaluate the methods mentioned in this paper on it. 

\section{Further Experiment on Confident Correspondence Generation}
%This appendix contains some further experiments. 

%\subsection{Confident correspondences in adaptive feature fusion}
As introduced in Section~\ref{sect:ff}, an important step in our adaptive feature fusion strategy is to detect confident correspondence. 
We report the percentage of correctly generated confident correspondences in Table~\ref{tab:confi}. 
It shows that our proposed confident correspondence generation strategy indeed leads to a high percentage of correct correspondences.

\begin{table}[htbp]
	\centering
	\caption{The percentage of correctly generated confident correspondences (PoC).}
	\resizebox{0.8\textwidth}{!}{
		\begin{tabular}{ccccccccc}
			\toprule
			\multirow{2}[3]{*}{} & \multicolumn{3}{c}{\dbps} & \multicolumn{4}{c}{\srp}     & \multirow{2}[3]{*}{\fb} \\
			\cmidrule{2-8}          & ZH-EN & JA-EN & FR-EN & EN-FR & EN-DE & DBP-WD & DBP-YG &  \\
			\midrule
			PoC   & 85.6\% & 87.1\% & 90.1\% & 89.0\% & 93.5\% & 93.3\% & 93.3\% & 82.6\% \\
			\bottomrule
		\end{tabular}%
	}
	\label{tab:confi}%
\end{table}%

%\subsection{Quicksort}

\end{document}